\documentclass[letterpaper, 10 pt, conference]{ieeeconf}  

\IEEEoverridecommandlockouts                              

\usepackage{graphicx} 
\usepackage{mathptmx} 
\usepackage{times} 
\usepackage{amsmath} 
\usepackage{amssymb}  
\usepackage{subfig}
\usepackage{underscore}
\graphicspath{{figures/}}
\newtheorem{definition}{Definition}
\newtheorem{proposition}{Proposition}
\DeclareMathOperator*{\argmin}{arg\,min}

\newcommand{\comment}[1]{#1}

\title{\LARGE \bf
Clinical Patient Tracking in the Presence of Transient and Permanent Occlusions via Geodesic Feature
}

\author{Kun Li$^{1}$ and Joel W. Burdick$^{1}$
\thanks{*This work was
supported by the National Institutes of Health, NIBIB.} \thanks{$^{1}$Kun Li
and Joel W. Burdick are with Department of Mechanical and Civil Engineering, California Institute of
Technology, Pasadena, CA 91125, USA {\tt\small kunli@caltech.edu}
}%
}

\begin{document}

\maketitle
\thispagestyle{empty}
\pagestyle{empty}

\begin{abstract}
  This paper develops a method to use RGB-D cameras to track the motions of a human spinal cord
  injury patient undergoing spinal stimulation and physical rehabilitation.  Because clinicians must
  remain close to the patient during training sessions, the patient is usually under permanent and
  transient occlusions due to the training equipment and the movements of the attending clinicians.
  These occlusions can significantly degrade the accuracy of existing human tracking methods. To
  improve the data association problem in these circumstances, we present a new global feature based
  on the geodesic distances of surface mesh points to a set of {\em anchor points}.  Transient
  occlusions are handled via a multi-hypothesis tracking framework. To evaluate the method, we
  simulated different occlusion sizes on a data set captured from a human in varying movement patterns,
  and compared the proposed feature with other tracking methods. The results show that the proposed
  method achieves robustness to both surface deformations and transient occlusions.  
\end{abstract}

\section{Introduction}

Approximately 400,000 Americans suffer from a severe spinal cord injury (SCI), which limits or
eliminates their ability to initiate and control voluntary motions. Recently, epidural spinal
stimulation has shown promise in helping to restore motor and autonomic function after major SCI
\cite{tracking:lancet}. To obtain the best results, spinal stimulation must be combined with
physical therapy, such as stand training, wherein a spinally stimulated patient with SCI attempts to
maintain an upright posture in a standing frame.  This training challenges the spinal cord control
circuitry, helping it to relearn function in the presence of spinal stimulation.  Particularly in
the early stand training stages, therapists assist the patient with the standing process, and
provide valuable sensory cues to the nervous system through physical contact.

To assess the patient's therapy progress, and to provide data which supports progress in research on
spinal stimulation therapy, the patient's motions during these training epochs must be accurately
tracked. Current clinical practice uses marker-based systems, as as the Vicon system.

However, patients should be able to stand-train in their home environment, which precludes the use
of high-cost marker-based systems. Moreover, clinics sited in emerging countries typically are not
equipped with marker-based systems. Hence, there is a need for low cost methods to precisely track
SCI patient motions during spinally stimulated stand and step training. Moreover, due to the
complicated occlusions described below, even marker-based tracking systems require extensive
hands-on manipulation of the tracking data after the training session in order to obtain useful
results.  Hence, our goal is to enable robust, accurate, and highly automated tracking of human
posture and motion using low-cost RGB-D camera technology in clinical environments where there are
many permanent and transient occlusions.  This paper focuses on off-line evaluation of the patient's
motion, as a first step toward a future real-time system.

\comment{
\begin{figure}
\centering
\includegraphics[width=7cm]{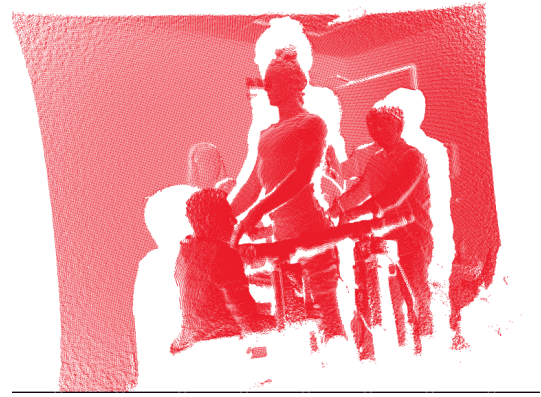}
\caption{\small Depth data captured by a Kinect V2 in a clinic during a rehabilitation session. The
  patient (middle) stands within a standing frame, aided by two therapists.  The left therapist
  holds the patient's knees, while the right one holds the patient's hips. The standing frame
  constantly and permanently occludes parts of the patient's body.  The therapists introduce
transient occlusions.}
\label{fig:ExampleOcclusions}
\vskip -0.25 true in
\end{figure}
}

As reviewed below, the use of low cost RGB-D sensors (such as the Microsoft Kinect or Intel
RealSense cameras) to track human motion has been widely investigated. However, two characteristics
of our application have not been well addressed in previous work. First, multiple therapists work in
very close proximity to the patient, often touching them for extended periods of time. The presence
and movements of these other people represent time varying and unpredictable
occlusions. Additionally, the rehabilitation equipment creates occlusions which are fixed in space,
but whose relationship to the tracked patient changes due to their movement.

\begin{figure}
\centering
\includegraphics[width=0.3\textwidth]{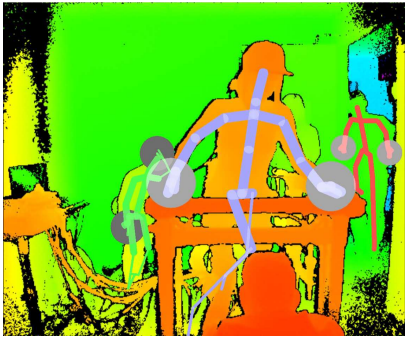}
\caption{\small Example of a Kinect V2 tracking a patient in a stand frame with three attending
  therapists: the sensor captures the front view of a patient (middle), who stands in the frame,
  with other therapists in front and behind. The skeletons are generated by the Kinect's 
  model-based tracking algorithm, and they are distorted by occlusions and distractions.}
\label{fig:modelsolution}
\vskip -0.25 true in
\end{figure}

In our experience, most human tracking algorithms are negatively affected by the occlusions,
especially the transient occlusions resulting from clinician movements (e.g., see Figure
\ref{fig:modelsolution}). The close proximity of other humans presents many challenges to appearance
based models.

This paper presents a new methodology to track humans in these complexly occluded conditions. We
represent the human body via a surface mesh. Metrics of clinical interest, such as joint angles, can
be derived from the mesh.  To enable accurate data association in the presence of time varying
occlusions, we design a novel surface mesh feature descriptor based on the geodesic distance between
mesh nodes (Section \ref{tracking:feature}). This feature is useful in the case of moderate surface
deformations between frames and the presence of feature-poor homogeneous surface regions.  Transient
occlusions are handled by a multi-hypothesis tracking framework \cite{tracking:mht} (Section
\ref{tracking:vertextracking}).  Experiments (Section \ref{tracking:experiments}) validate the
method and contrast its performance and robustness with respect to several methods found in the
recent literature.

\section{Related Work} \label{tracking:relatedworks}

Because of its importance to many applications, the subject of human pose estimation and body motion
tracking has received considerable attention.  For pose estimation, sensory data is often organized
as a surface mesh or skeleton.  E.g., Anguelov et. al \cite{tracking:scape} construct a mesh with a
largest set of markers on the human body, and describe changes in pose based on mesh deformations.
We pursue a markerless strategy due to its greater convenience. Combined with a learning process,
the pose model can be estimated from markerless data. Dantone et. al \cite{tracking:part} adopt a
part-based human model and use a random forest to train the model. Shotten et. al
\cite{tracking:kinect} train a random forest on a large set of body shape and pose data, with a
simple feature to classify each body point. The classification is used to predict the joint
locations.  Unfortunately, these approaches can fail when assisting therapists are in close
proximity to the patient.

Clinicians often wish to track selected points on the human body.  In a markerless system, the
tracking algorithm must track the selected points across frames. In 2D images, these features are
usually based on image edges and gradients, like the histogram of shape context \cite{tracking:2dsc}
and histogram of oriented gradient \cite{tracking:2dhog}. These features can be extended to
three-dimensional space by describing each point in a point cloud, like fast point feature histogram
\cite{tracking:fpfh}, signature of histograms of orientations \cite{tracking:shot} and shape context
\cite{tracking:shape3d}, which describe each point based on local appearance and local curvature.
Our clinical application introduces permanent and transient occlusions in two ways. First, some body
parts or selected points may be invisible in certain frames, leading to missing tracking
results. Second, even if the selected points are visible, their feature descriptors may be
erroneous, since they can depend on occluded points.

Data can be merged from multiple views to reduce permanent occlusion problems.  E.g., Zhang et. al
\cite{tracking:multicamera} merge multiple point clouds, and then track human pose with a particle
filter. Dockstader and Tekalp \cite{tracking:mutlicamera1} use a Bayesian belief network to merge
individually processed depth frames. We minimize the effect of permanent occlusions with multiple
camera views and an apriori model of the training equipment.

The effects of transient occlusions can be reduced using local surface mesh features which only
change slightly under deformations.  MeshHOG \cite{tracking:meshfeature} samples and triangulates a
surface mesh uniformly to extract local geometric and photometric properties. LD-SIFT
\cite{tracking:meshfeature1} extends the SIFT image descriptor to 3D meshes.  These local features
may be invariant to transient occlusions and geodesic mapping, but in clinical tracking they yield
inaccurate matchings due to homogeneous regions on the human body.

Another category of mesh-based local features is based on spectral shape analysis, such as the heat
kernel signature \cite{tracking:hks}, scale-invariant heat kernel signature \cite{tracking:sihks},
and wavelet kernel signature \cite{tracking:wks}.  Theoretically, these features are robust to
occlusions, deformations, and homogeneous regions, but we observe many inaccurate matchings in
practice.

We design a global surface feature robust to deformations and homogeneous regions, based on
generalized multidimensional scaling \cite{tracking:GMDS}, and then use the framework of multi
hypothesis tracking \cite{tracking:mht} to handle the transient occlusions.

\section{Problem Formulation}
\label{tracking:problem}

Patients undergo spinal stimulation while executing a sequence of rehabilitation or testing moves
with the help of clinicians and a rehabilitation device.  We assume that multiple RGB-D sensors
\footnote{The optimal placement of the RGB-D sensors is beyond the scope of this paper.  We assume
  that if the rehabilitation devices and the clinicians were not present, the cameras could capture
a full frontal view of the patient.  Moreover, we assume synchronized shutters.} capture the
subject's movements from different perspectives. By attaching a coordinate system $\Gamma_0$ to the
training device, we build a 3D model, $\chi_0$, of the device in $\Gamma_0$ based on manual measurement.

\begin{figure}
\centering
\subfloat[Front view of the subject]{
\includegraphics[width=0.15\textwidth]{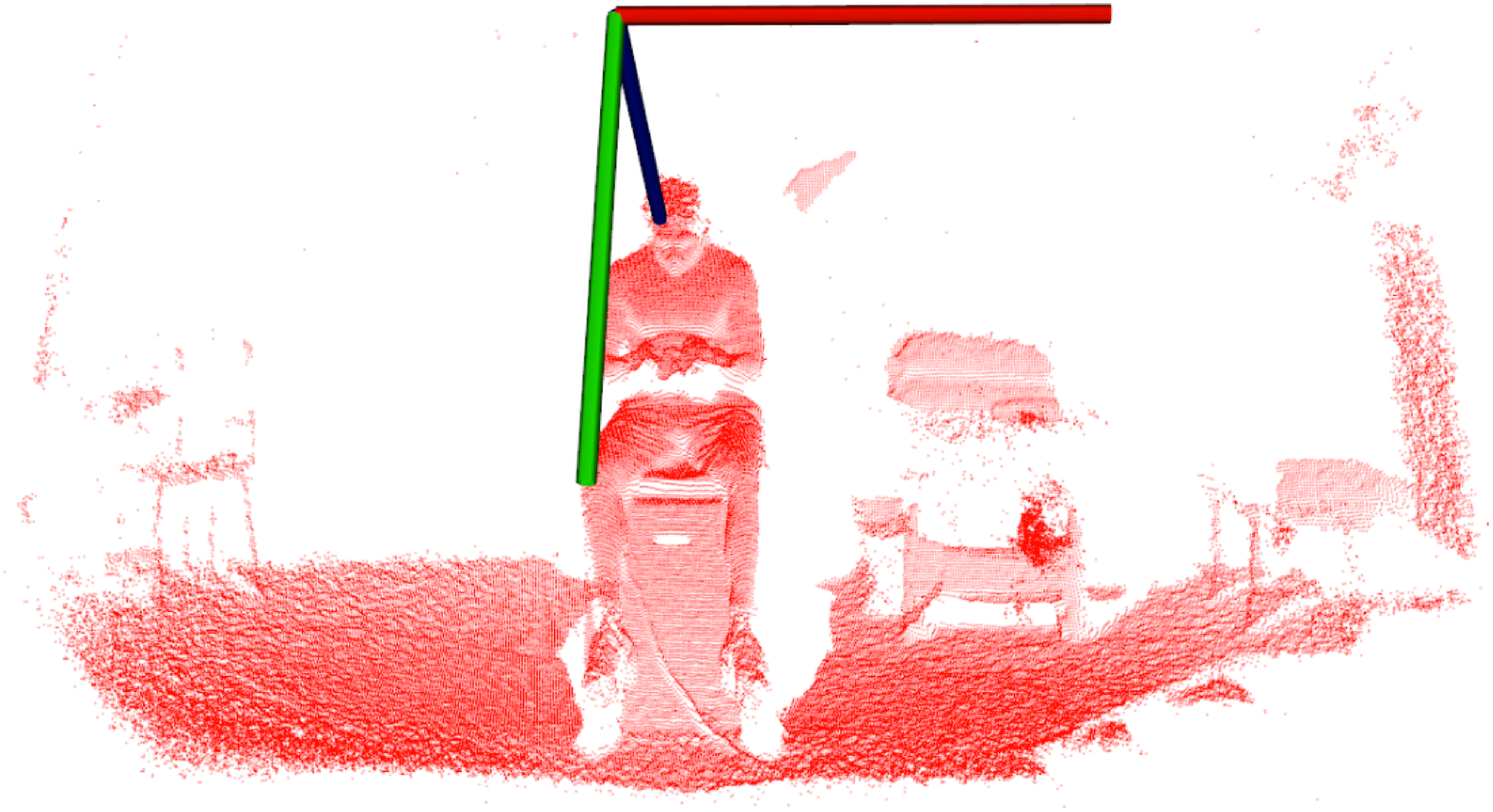}
}
\qquad
\subfloat[Side view of the subject]{
\includegraphics[width=0.15\textwidth]{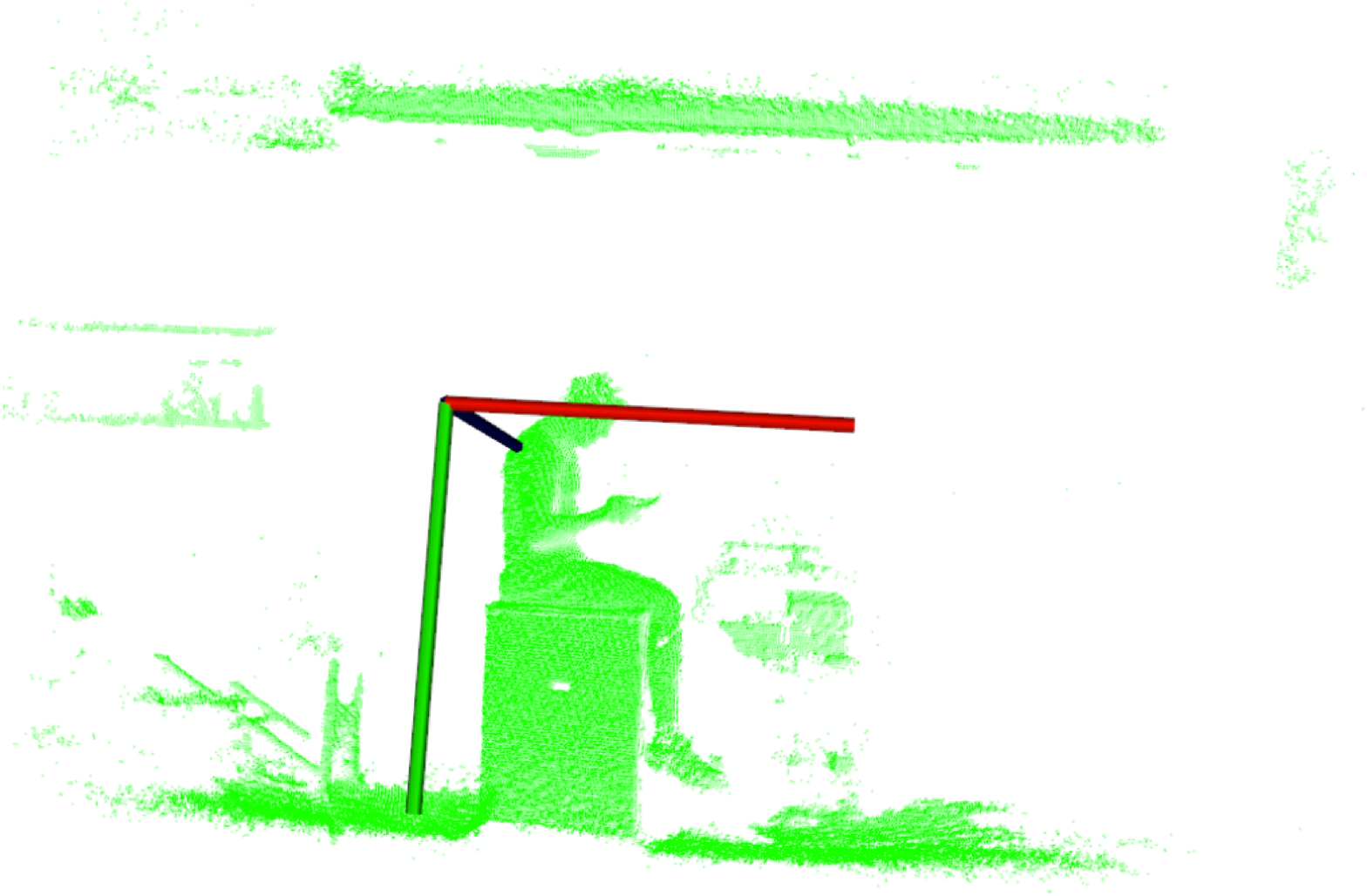}
}
\\
\subfloat[Synthesized data from two cameras]{
\includegraphics[width=0.15\textwidth]{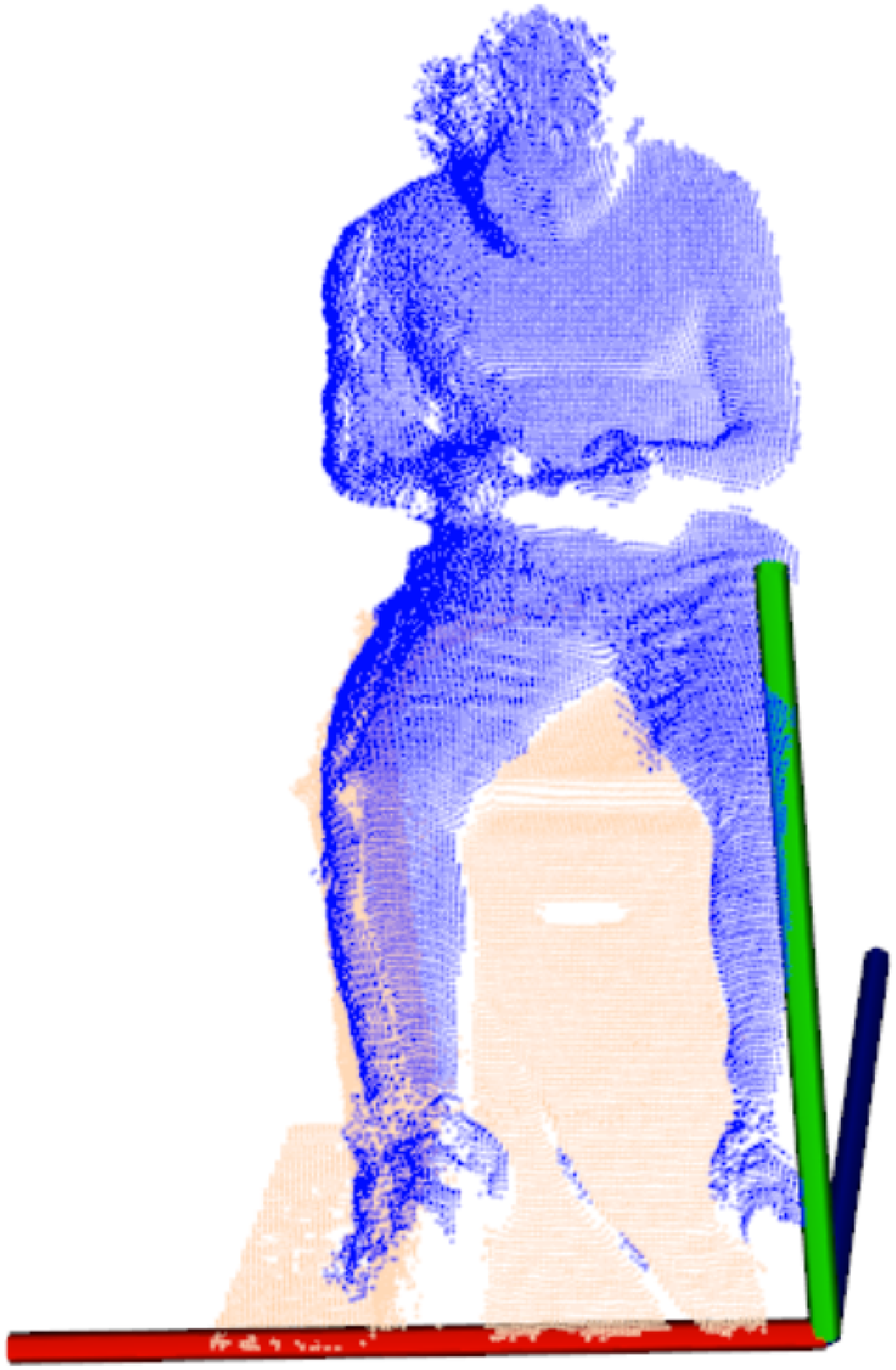}
}
\qquad
\subfloat[Data of the isolated subject]{
\includegraphics[width=0.15\textwidth]{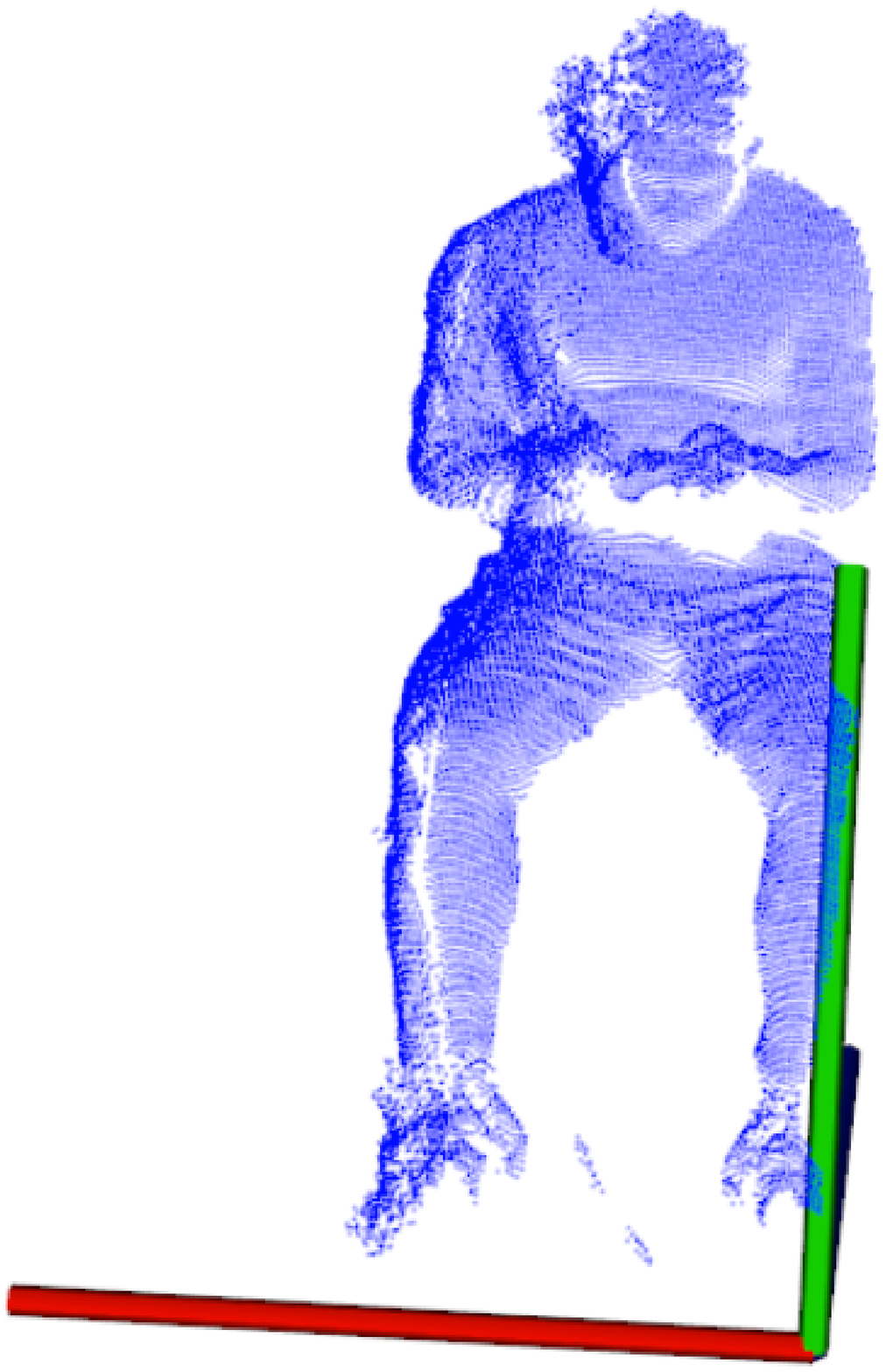}
}
\caption{\small Calibration and removal of permanent occlusions}
\label{track::calibration}
\vskip -0.25 true in
\end{figure}

Numbering the sensors as $\{1,\cdots,K\}$ and their coordinate systems as $\{\Gamma_1, \cdots,
\Gamma_K\}$, we detect the training device pose in each camera to estimate a set of transformation
matrices, $\{\tau_{0,i}\}$, between $\Gamma_0$ and $\{\Gamma_1,\cdots,\Gamma_K\}$:
    \begin{equation}\{\tau_{0,i}|i=1,\cdots,K\}\end{equation}
where $\tau_{0,i}$ is the transform between $\Gamma_0$ and $\Gamma_i$.  Using these device-to-sensor
transforms, the range data from each camera is aligned into the common coordinate system $\Gamma_0$.
Most of the range data associated with the permanent occlusions, such as the training device, can be
removed using the prior model $\chi_0$. Since patient moves are limited within the training device,
range-based background subtraction isolates the sensory data, $L$, associated to the patient(see
Fig.  \ref{track::calibration}).  Many therapist body parts may be included in $L$ after this process.

The processed RGB-D data points at time $t$ are first converted into a surface mesh, $M^t$, via a
triangulation algorithm \cite{tracking:triangulation} implemented in Point Cloud Library
\cite{tracking:PCL}: 
\begin{equation}  M^t=\{V^t,E^t\} \end{equation} 
where $V^t=\{v_i^t\}$, denotes the nodes of the mesh, and $E^t=\{e_{ij}^t\}$, denotes the edges
between adjacent mesh nodes at frame $t$. This algorithm uses a Weighted Least Squares method to
compute surface normal, thus it is robust to noises in the RGB-D data. Besides, we do not assume
that the mesh contains the same number of nodes across successive frames.

\comment{We adopt a classical probabilistic filtering framework for the algorithm architecture.  The
  filter incorporates: (1) a prediction step where a dynamic motion model predicts the system state
  at time $t+1$ given a state estimate at time $t$; (2) a data association step to match
  measurements at time $t+1$ with specific targets; and (3) a measurement update step to improve the
  state estimate at $t+1$ using the data association and the dynamic prediction.  We used a {\em
    scaled dynamics} approach \cite{tracking:smd} for the dynamic prediction step.  One of the main
  contributions of this paper is a new feature descriptor to improve the data associate process for
  dense depth data. }

We need an association function, $\phi^{t,t+1}$, that describes the correspondences between
the nodes of $M^t$ and $M^{t+1}$:
   \begin{equation} \phi^{t,t+1}: V^t\rightarrow V^{t+1}\ . \end{equation}
The function $\phi^{t,t+1}$ maps each node in $M^t$ to its associated node in $M^{t+1}$.  Due to
occlusions and noise, some nodes in $M^{t}$ have no associating nodes in $M^{t+1}$, and 
some nodes in $M^{t+1}$ have no predecessor nodes in $M^{t}$.

The association of mesh points across frames also allows us to track an articulated model of
the body, consisting of rigid bodies connected by joints.  Spherical joints (e.g. shoulder and hip
joints) are modeled by the point located at the joint center, while revolute joints (e.g., elbow or
knee) are modeled as a point and a rotation axis passing through that point. Different clinical
study objectives will dictate different types of models.  One can assign a rule, $J \colon M^{t_k}
\rightarrow P^{t_k}$, for defining the joint parameters, $P^{t_k}$, from the mesh geometry.
A set joint variables, which define the articulated model state, can be defined in terms of the
joint parameters and mesh nodes:
  \begin{equation}
    \theta_i^t = \{ \mathcal{T}(M^{t},P_i^{t}) | i=1,\cdots,N_J \}
  \end{equation}
where $N_J$ is the number of articulated model joints (see Figure \ref{tracking::update} for an
example). For our experiments, we use $N_J=21$ and a model which
describes each joint as a node separating two adjacent body parts, which themselves are defined
by a set of mesh nodes\footnote{In practice, the hand tip joints and thumb joints are not included, because
they are usually folded during rehabilitation.}.

\comment{The correctly associated mesh data points $M^{t+1}$ can be used to update the articulated
  model state estimate, $\hat{\theta}_{t+1}$, in two ways.  First, the nodes of $M^{t+1}$ can be
  associated to the states $\theta_{t+1}$ through a measurement function, $h(\theta_{t_1})$. This
  function allows for data assimilation in the measurement update step of an EKF, UKF, or particle
  tracking filter.  Alternatively, one can consider the mesh coordinates as the model state, and the
  articulated model states are obtained by applying the functions $J$ and $\mathcal{T}$ described
  above. }

\comment{
\begin{figure}
\centering
\includegraphics[width=0.25\textwidth]{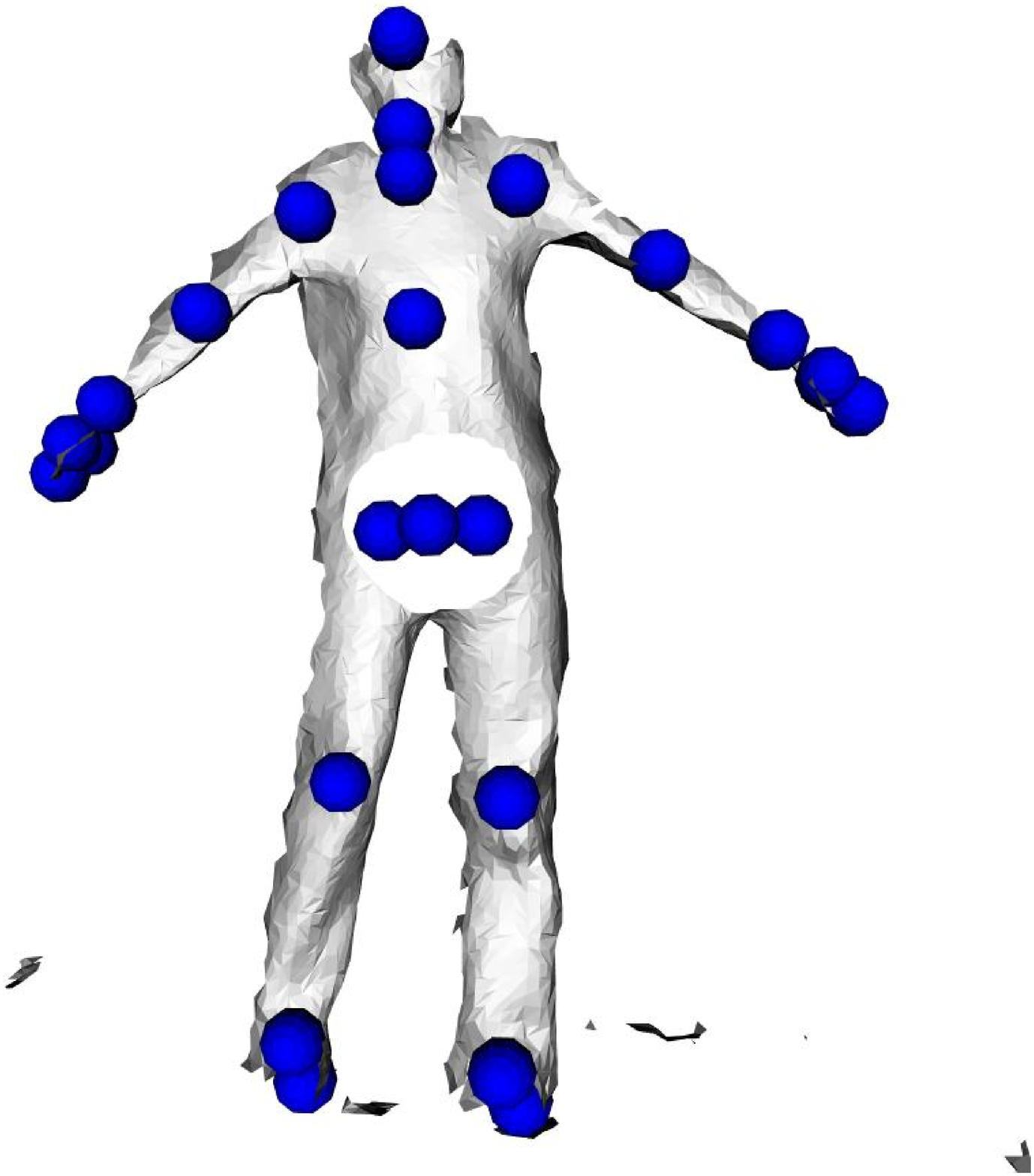}
\includegraphics[width=3.2cm]{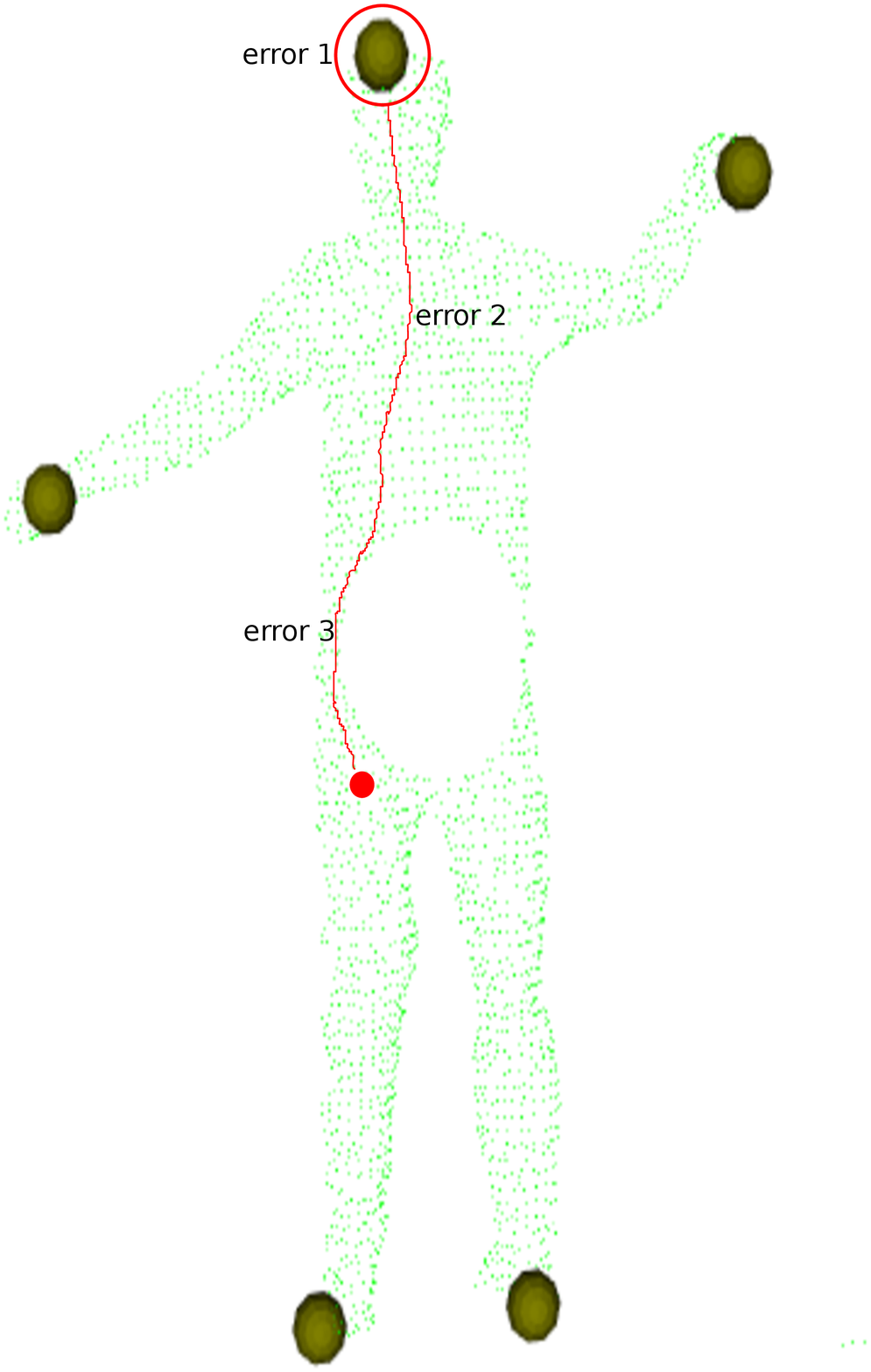}
\caption{\small {\bf Left:} An occluded surface mesh is created by triangulating a point cloud and removing
data points. The joint locations are detected by Kinect on the original unoccluded point cloud.
{\bf Right:}  The distance between the red vertex and the yellow anchor node is sensitive to 
errors: error 1 is anchor node localization error; error 2 is unmodeled surface changes
between the vertex and anchor; error 3 denotes geodesic distance changes due to transient occlusions.
}
\label{tracking::mesh}
\vskip -0.25 true in
\end{figure}
}

\section{The Geodesic Vertex Descriptor} \label{tracking:feature}

If $M^t$ is viewed as a 2D manifold, and if mesh distortions between frames are modest, they can be
described by a mapping in which the geodesic distance between every vertex pair is approximately
preserved.  I.e., the function $\phi^{t,t+1}$ defines an (approximately) isometric embedding of the
$M^t$ to $M^{t+1}$, and it can be estimated by generalized multi-dimensional scaling (GMDS)
\cite{tracking:GMDS}, which chooses the embedding that minimizes the distortion of geodesic
distances.

However, our clinical application challenges this idea.  A mesh patch in $M^t$ is unlikely to be an
exact subset of the corresponding patch in $M^{t+1}$.  Consequently, the distortion of geodesic distances
may be in accurate due to occlusions. To solve the first problem, we design a feature descriptor
inspired by GMDS, and approximate the feature with a set of anchoring nodes. To solve the second
problem, we adopt a multi-hypothesis tracking framework \cite{tracking:mht}.

\subsection{Generalized Multi-dimensional Scaling}

An ideal mapping $\phi^{t,t+1}$ should preserve the geodesic distance between every two nodes
$v,v^{'}\in M^t$ under the mapping to $\phi^{t,t+1}(v),\phi^{t,t+1}(v^{'})\in M^{t+1}$.  In
practice, $\phi^{t,t+1}$ is chosen to minimize the differences between geodesic distances.

Formally, for surfaces $M^t$ and $M^{t+1}$, a mapping function $\phi^{t,t+1}\colon V^t \rightarrow
V^{t+1}$ induces a mesh distortion
  \begin{equation}\label{eq:distortion}
     dis\, (\phi^{t,t+1}) = \sup_{v,v'\in V^t}|gd_{M^t}(v,v') -
        gd_{M^{t+1}}(\phi^{t,t+1}(v),\phi^{t,t+1}(v'))|
  \end{equation}
where $gd_{M^t}(x,x^{'})$ denotes geodesic distance between nodes $x,x^{'}\in M^t$ \cite{tracking:GMDS}.
GMDS computes the following distance to find the optimal correspondences between $M^t$ and $M^{t+1}$.
  \begin{equation}
    \label{eq:gmds}
      d_{PE}(M^t,M^{t+1})=\frac{1}{2}\inf_{\phi^{t,t+1}} dis\,\,\, \phi^{t,t+1}
  \end{equation}

Inspired by this distance, let us first consider a hypothetical feature descriptor for vertex $v$ in
mesh $M^t$ based on its ordered distances to all other vertices of $M^t$:

\begin{definition} \label{defn:feature}
Let $M^t$ be simply connected (it contains no holes, but it can have a boundary).  The
{\em unoccoluded geodesic distance feature descriptor}, $f^U_{M^t}(v)$, for node $v\in V^{t}$ is:
   \begin{equation} f^U_{M^t}(v) = \{gd_{M^t}(v,v')|v' \in V^t\})\ . \end{equation}
\end{definition}

As a small contribution, we note that GMDS is essentially equivalent to a matching scheme that
minimizes the Euclidean distance between the feature descriptors of two matched nodes:
  \begin{equation}
   \label{eq:euclidean}
     \phi^{t,t+1}=\argmin_{\phi^{t,t+1}}\sum_{v\in V^t}ed(f^U_{M^t}(v),f^U_{M^{t+1}}(\phi^{t,t+1}(v)))
  \end{equation}
where $ed(f^U_{M^t}(v),f^U_{M^{t+1}}(\phi^{t,t+1}(v)))$ is the Euclidean distance of two feature
descriptors.
\comment{The equivalence with GMDS can be seen as follows.  Denote the value of the
  distortion in (\ref{eq:distortion}) for a given $M^t$ and $M^{t+1}$ as $\epsilon$, then the sum of
  all of the distortions due to mapping function $\phi^{t,t+1}$ is:
\begin{align}
&\sum_{v \in V^t}ed(f_{M^t}(v),f_{M^{t+1}}(\phi^{t,t+1}(v)))\nonumber\\
&=\sum_{v \in V^t}\sqrt{\sum_{v' \in V^t}(gd_{M^t}(v,v')-gd_{M^{t+1}}(\phi^{t,t+1}(v),\phi^{t,t+1}(v')))^2}\nonumber\\
&=\sum_{v \in V^t}\sqrt{\sum_{v' \in V^t}|gd_{M^t}(v,v')-gd_{M^{t+1}}(\phi^{t,t+1}(v),\phi^{t,t+1}(v'))|^2}\nonumber\\
&\leq \sum_{v \in V^t}\sqrt{\sum_{v'\in V^t} \epsilon^2} =N\sqrt{N} \epsilon\nonumber
\end{align}
where $N$ denotes the number of vertices in $M^t$.  Therefore, Equation (\ref{eq:gmds}) minimizes
the upper bound of the Euclidean distance while Equation (\ref{eq:euclidean}) minimizes the Euclidean distance.
}

\subsection{Approximate Geodesic Feature Descriptor}

GMDS makes assumptions about meshes $M^{t}$ and $M^{t+1}$ which may not be true in practice.  GMDS
requires that $M^t$ is a model template: the nodes of mesh $M^{t+1}$ must be a subset of the nodes
of $M^t$.  This assumption does not hold in our application, since patient movements and occlusions
may cause missing or newly emerging vertices, leading to inaccurate feature descriptors.  To solve
the problem, we propose an approximation to $f^U_{M^t}(v)$ consisting of a small set of {\em anchor
  nodes}, $A$, that are stably present in both meshes.
\begin{definition}\label{defn:AnchorNode}
  An {\bf anchor node}, $A_i$, is a node of the surface mesh which has two characteristics:
  \begin{enumerate}
  \item{} $A_i\in V^t$ and $A_i\in V^{t+1}$,
  \item{} $A_i$ is identifiable in meshes $M^t$ and $M^{t+1}$,
  \end{enumerate}
\end{definition}
As described below, anchor nodes lead to a new and practically useful feature descriptor which
measures the geodesic distances between a given mesh node and the set of anchors.  In human mesh
matching, we use local extremas (e.g., hands, feet, and head of the patient) as the anchors:
     \begin{equation}A^t=\{A_j^t|j=1,\cdots,N_A\}\end{equation}
where each $A_j$ denotes the $j^{th}$ anchor node, and $N_A$ denotes the number of anchor nodes.

To locate the anchors we define the center of $M^t$, $V^t_c$, as the average position of all
mesh points.  Anchor nodes have a larger geodesic distance to $V^{t_c}$ than their neighbors:
  \begin{equation}
     A^t=\{v|v \in V^t, \forall v_n\in N(v), gd_{M^t}(v^t_c,v)>gd_{M^t}(v^t_c,v_n)\}
   \end{equation}
where $N(v)=\{v_n|v_n \in V^t, gd_{M^t}(v_n,v)<r\}$ denotes the set of vertices within radius
$r$ from $v$, measured by geodesic distance. The radius is empirically selected. These
nodes are the geodesic extremas \cite{tracking:geodesic}, and they are ordered based on the corresponding body parts.
Using the anchor nodes, each vertex $v^t_i$ in mesh $M^t$ is described with the following feature.
\begin{definition}\label{defn:AnchorFeature}
  Let $A^t$ be the set anchor nodes in $M^t$.  The {\bf anchor-based feature}, $f^t_i$, of 
  $v^t_i\in M^t$ is an $N_A$-tuple whose elements are the geodesic distances from
  $v^t_i$ to each anchor.
\begin{equation}
  f^t_i = \{gd_{M^t}(v^t_i,A^t_j)|j=1,\cdots N_A\}
\end{equation}
\end{definition}

\comment{
\subsection{Error Analysis}

Three errors will reduce the accuracy of feature descriptors based on geodesic distance: (1) errors
in localizing the anchors, (2) unmodeled surface mesh deformations, and (3) occlusions, as shown in
Figure \ref{tracking::mesh}(right).  Hence, the measured geodesic distance between $v^t_i$ and anchor
$A^t_j$ is:
  \begin{equation}
    gd_{M^t}(v^t_i,A^t_j) = \hat{gd_{M^t}}(v^t_i, A^t_j) + \xi(A^t_j)+\epsilon(v^t_i,A^t_j)
                           + \sigma(v^t_i,A^t_j)
  \end{equation}
where $\hat{gd_{M^t}}(v^t_i,A^t_j)$ denotes the true geodesic distance from $v^t_i$ to $A^t_j$,
$\xi(A^t_j)$ denotes the $j^{th}$ anchor node localization error, $\epsilon(v^t_i,A^t_j)$ denotes
the error due to unmodeled sensor noise, clothes deformations, and underlying musculature, and
$\sigma(v^t_i,A^t_j)$ denotes the geodesic path distortion introduced by the presence of
occlusions between $v^t_i$ and $A^t_j$.

The Euclidean distance between the feature vectors of a node $v^t_i$ and its corresponding node in 
$M^{t+1}$ is:
  \begin{multline}
    ed(f^t_i,f^{t+1}_k) = \sum_{j=1}^{N_A}(\hat{gd_{M^t}}(v^t_i,A^t_j) -
      \hat{gd_{M^{t+1}}}(v^{t+1}_k,A^{t+1}_j) \\ + e_1 + e_2 + e_3)^2
\end{multline}
where $e_1=\xi(A^t_j)-\xi(A^{t+1}_j)$, $e_2=\epsilon(v^t_i,A^t_j)-\epsilon(v^{t+1}_k,A^{t+1}_j)$,
and $e_3=\sigma(v^t_i,A^t_j)-\sigma(v^{t+1}_k,A^{t+1}_j)$. Minimizing these these errors will
improve the data association process.

The error $e_1$ captures changes in the geodesic extrema due to noise, and is bounded for bounded
sensor noise.  Similarly, $e_2$ is bounded since body and cloth distortions between frames are
limited in practice. Hence, we can identify a upper limit on $e_{max}=e_1+e_2$, and find all nodes
within that limit:
\begin{equation}
      \psi_{V^{t+1}}(v_i^t)=\{v|v\in V^{t+1},||f^t(v^t_i)-f^{t+1}(v)||^2<e_{max})\}
  \end{equation}
All the nodes in $\psi_{V^{t+1}}(v_i^t)$ are possible matching vertices of $v^t_i$ in $V^{t+1}$,
if error $e_3$ can be excluded.  We choose a node closest to their average position as the most
probable match:
  \begin{equation} v^{t+1}_k=avg(\psi_{V^{t+1}}(v^t_i)) \end{equation}
This solution is accurate only when the occlusion error $e_3$ can be excluded.  We solve this
problem with multi-hypothesis tracking: we generate a set of data matching hypothesises and find the
one least affected by error term $e_3$.  }

\comment{
\subsection{Properties of the Feature Descriptor} \label{sec:properties}

For good and robust tracking, a mesh point feature descriptor should be unique and accurate.  When
all mesh nodes are anchors, uniqueness is guaranteed unless the mesh is highly symmetric. The
minimum number of anchor nodes needed to ensure uniqueness is generally unknowable.  Our experience
shows that, when occlusions are not present, three anchor nodes suffice.  Due to the preservation of
geodesic distance under deformations, a point feature descriptor will not change during pose
changes.  If uniqueness holds in one pose, it holds for all poses. Assume that the initial body pose
has no self-occlusion, and it is mapped to the Euclidean plane via a bijection. In the Euclidean
plane, it is trivial to show that three anchor nodes ensures uniqueness of a point.

The geodesic feature accuracy can degrade under occlusions since geodesic distance to every anchor
is not preserved.  The preservation of geodesic distance to the anchors depends on the relative
anchor locations, and the occlusion size and shape.  Next we develop some results on distance
preservation with limited convex occlusions.  The proofs are eliminated due to length restrictions.

\begin{definition} \label{defn:bright}
Consider a convex occlusion, $O$, in mesh $M$. For  $v_i\ \in\ M$, the {\bf bright nodes} of $v_i$
are the set of nodes in the boundary of $O$ whose geodesic distances to $v_i$ are undistorted. The
{\bf bright edges} of $v_i$ are the edges in the boundary of $O$ which connect the bright nodes.
The {\bf bright region} of $v_i$ is the set of bright nodes and edges.
\end{definition}

\begin{definition} \label{defn:shadow}
For occlusion $O$ in mesh $M$ the {\bf shadow nodes} of $v_i\in M$ are the nodes whose geodesic
distance to $v_i$ are distorted by the presence of the occlusion.  A {\bf shadow edge} is an edge in
$M$ that connects two shadow nodes.  The {\bf shadowed occlusion boundary} are those shadow nodes
and edges which lie on the boundary of $O$.
\end{definition}

These ideas are illustrated in Figure \ref{track::obstacleproof1}.
\begin{figure}
\centerline{
\includegraphics[width=0.3\textwidth]{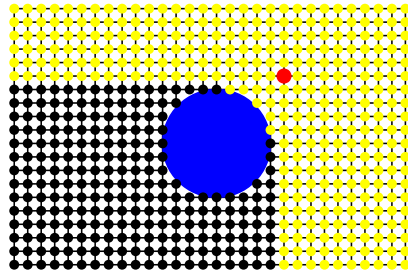}
}
\caption{\small Depiction of bright and shadow regions: the shadow region of the red node is plotted
as black nodes, and the bright region is plotted as in yellow nodes.}
\label{track::obstacleproof1}
\vskip -0.2 true in
\end{figure}

\begin{proposition}
If the bright regions of two nodes, $v_1$, $v_2$, on occlusion $O$ intersect, then $gd(v_1,v_2)$
must be undistorted by the presence of $O$.
\end{proposition}

This result can be extended to the following.

\begin{proposition}
If the border of occlusion $O$ is covered by the union of all anchor point bright regions
$B_A=\{B_{A_1}\cap\cdots\cap B_{A_{N_A}}\}$, then every mesh node  will have an undistorted
geodesic distance to at least one anchor node.
\end{proposition}

In the presence of multiple occlusions, some mesh points will experience distorted geodesic
distances to all anchoring nodes. A multiple hypothesis testing method, discussed next, helps to
alleviate the errors arising in such situations.
}

\section{MHT Vertex Tracking} \label{tracking:vertextracking}

We add a multi-hypothesis-tracing (MHT) framework to further improve data association across frames
in the presence of multiple occlusions. MHT requires a gate function, $\mathcal{G}$, to generate a
set of hypotheses, $H^{t+1}_{v^t_i}$, which describe the possible correspondences between vertex
$v^t_i\in M^t$, and vertices in in $M^{t+1}$:
$H^{t+1}_{v^t_i}\ =\ \mathcal{G}(f^t_i,M^t,M^{t+1})$. From the gate function, we construct a
tracking tree $Tr^{t:N_{TR}}$ which links nodes in $M^t$ to their hypothesized children nodes in
$M^{t+1},\ldots,M^{t+N_{Tr}}$, where $N_{Tr}$ denotes the tree depth.  Each branch of
$Tr^{t:N_{TR}}$ is scored, as described below.  The scored tree is searched for an optimal
association hypothesis.

\subsubsection{Track Tree Construction}

We assume that not all geodesic distances between vertices and anchors are corrupted by
occlusions.  Since we do not know which subset of distances $S_A$ is unaffected, we generate a set
of feature descriptor hypotheses for each node $v^t_i$, where each hypothesis is based on one
possible subset of anchoring nodes.

Based on this idea, we generate a set of possible feature descriptors for vertex $v^t_i$:
\begin{equation}f^t_{i,m}=\{gd(v^t_i,A^t_j)|A^t_j \in S_{A^t,m},j=1,\cdots,N_{S_A}\}\end{equation}
where $m=1,\cdots,\binom{N_A}{N_{S_A}}$ denotes the index of each possible subset, $N_{S_A}$ denotes
the number of anchor nodes in $S_A$, and $S_{A^t,m}$ denotes one possible subset of nodes.

\begin{figure}[h]
\centering
\includegraphics[width=0.35\textwidth]{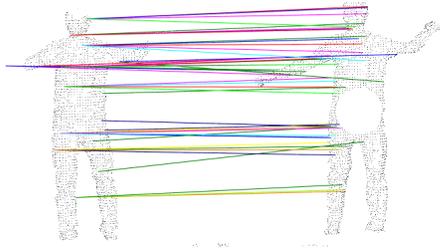}
\caption{\small Possible matches between points in the left point cloud and points in 
the right cloud are plotted, with color denoting a hypothesis.}
\label{tracking::hypothesis}
\vskip -0.25 true in
\end{figure}

Each descriptor leads to one hypothetical association of the nodes between two frames, thus for each
node $v^t_i$, we have $\binom{N_A}{N_{S_A}}$ possible matched nodes in frame $M^{t+1}$. An example
of the possible matches is shown in Figure \ref{tracking::hypothesis}.

Given these hypothesis, we build a track tree $Tr^{t:N_{Tr}}$ by linking nodes to their
$\binom{N_A}{N_{S_A}}$ children in next frame.

\subsubsection{Track Tree Scoring}
We score each tree branch to find the best node association hypothesis. 
Branch $Tr^{t:N_{TR}}_{i,b}$ is scored on the distance between the measured and predicted node locations
    \begin{equation}R^{t:N_{TR}}_{i,b}=ed(Tr^{t:N_{TR}}_{i,b},P_{Tr^{t:N_{TR}}_i})\end{equation}
where $Tr^{t:N_{TR}}_{i,b}$ denotes the $b_{th}$ tree branch starting from node $v_i^t$, and
 $P_{Tr^{t:N_{TR}}_i}$ denotes the predicted node locations starting from node $v_i^t$.  Prediction of the node location in an MHT
framework requires a dynamic model to predict movement forward in time.

Because human motion is complex to model, we use a {\em scaled motion dynamics} \cite{tracking:smd}
empirical approach to predict node locations. However, any other procedure which provides useful
dynamic motion predictions can be used instead.  

To construct the scaled motion dynamics model, a reference data set, $D$, of human motions is 
scaled by interpolation and re-sampling to describe varying speed motion patterns:
   \begin{equation}D_s=\{\theta_s^t|t=1,\cdots,s*N_D\}\end{equation}
where $s$ denotes the scaling level, and $N_D$ denotes the number of samples in the reference set.
Note that patients undergoing spinal stimulation are asked to carry out highly stereotyped motions,
which are well suited to such a modeling approach. 

The labeled samples $\theta^{t-k},\cdots,\theta^{t-1}$ are compared with the training data to find
the best matched sub-list:
  \begin{equation}\hat{s},\hat{j}=\argmin_{s,j}\sum_{v=1}^{k}ed(\theta^{t-v}-\theta_s^{j-v}). \end{equation}
That is, the current estimate of body motion is used to search the motion data base for similar
motions. The most likely motion match is then used to predict the next model state:
  \begin{equation}\hat{\theta}^{t+1}
    =\theta^{t}+\theta_{\hat{s}}^{\hat{j}+1}-\theta_{\hat{s}}^{\hat{j}}. \end{equation}

Under each hypothesis, a tracked vertex will be occupy a series of locations in subsequent frames.
If the hypothesis is not affected by occlusions, these locations will be similar to the dynamic
predictions. Otherwise, the hypothesis is likely affected by transient occlusions, and these
locations will deviate from the predicted movement patterns.

Then we estimate a set of transformation matrices $\tau^{t,t+1}$ for all body parts based on the
prediction, and then transform each vertex $v^t_i$ into a predicted location $v^{t+1}_j$.  Repeating
this process on the next $N_{Tr}$ frames generates a predicted tree $P_{Tr^t_i}$ for each
$v^t_i$, which is used to score the tree branches. We restrict rapid tree growth by pruning
branches with low scores when the tree size exceeds an empirical threshold.
\subsubsection{Pose Estimation}

We search the scored tracking tree for the best hypothetical matching node pairs between frame $t$
and $t+1$, and use these pairs to estimate the joint locations in frame $t+1$. We construct the
global hypothesis by identifying the branch with highest score for each node and extracting the
matching node pairs from this branch.

\begin{figure}
\centering
\includegraphics[width=0.2\textwidth]{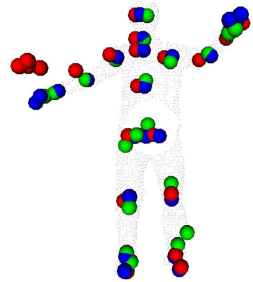}
\vskip -0.1 true in
\caption{\small Joint locations: the red points denote the predicted joint locations, the green points 
denote the updated joint locations, and the blue points denote the ground truth.}
\label{tracking::update}
\vskip -0.18 true in
\end{figure}

Given the matching node pairs, we use iterative closest points to estimate the transformation matrix
for each body part, with the predicted joint locations as initial guess:
\begin{equation}\hat{\tau}^{t,t+1}=ICP(\tau^{t,t+1},\phi^{t,t+1})\end{equation}
The joint locations can be extracted from the updated transformation matrices $\hat{\tau}^{t,t+1}$.
One example of the predicted and updated joint locations is shown in Figure \ref{tracking::update}.

\section{Experiments} \label{tracking:experiments}

\begin{figure}
\centering
\subfloat[Matching  with fast point feature histogram]{
\includegraphics[width=0.15\textwidth]{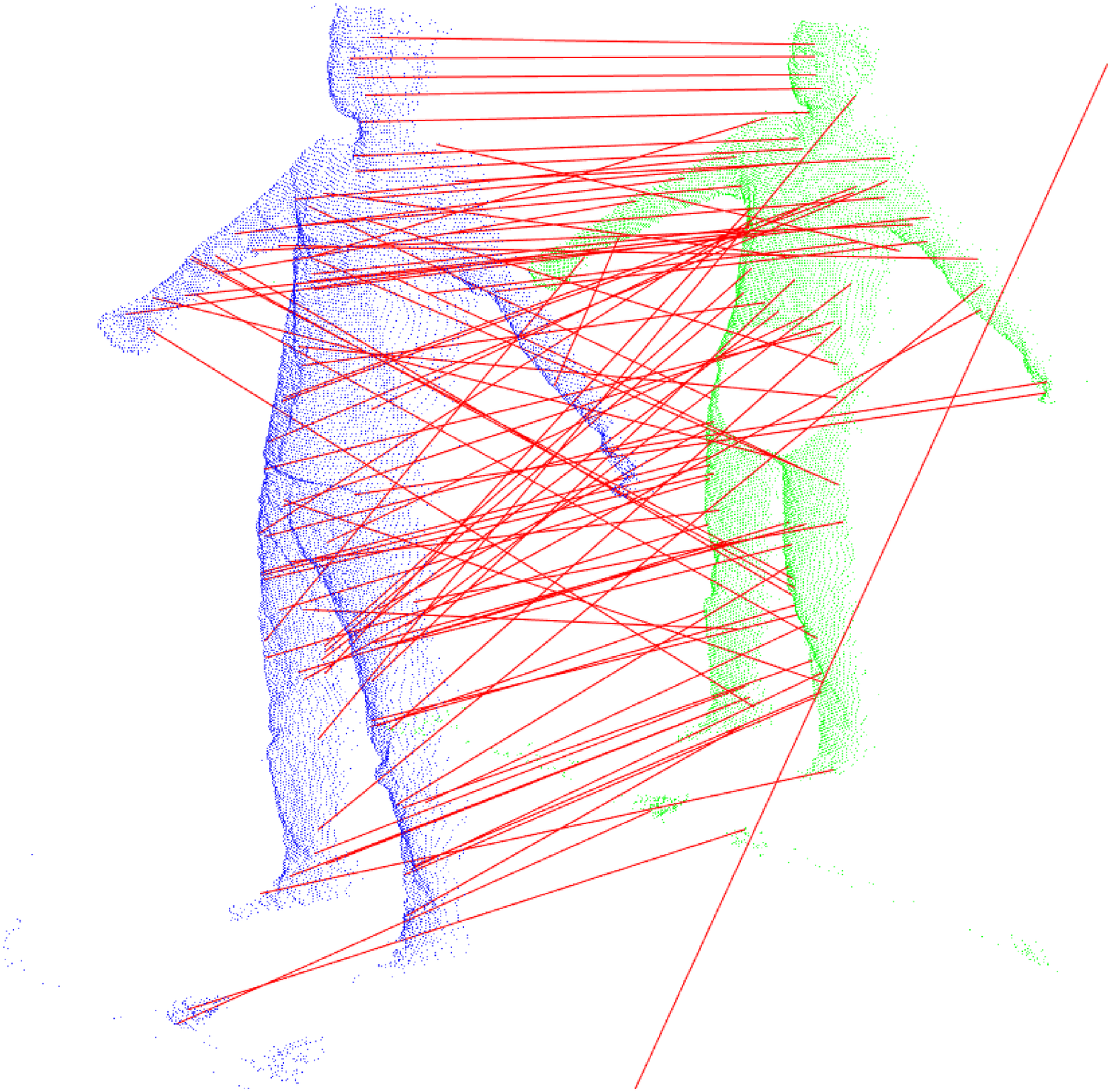}
}
\qquad
\subfloat[Matching with rotation projection statistics]{
\includegraphics[width=0.15\textwidth]{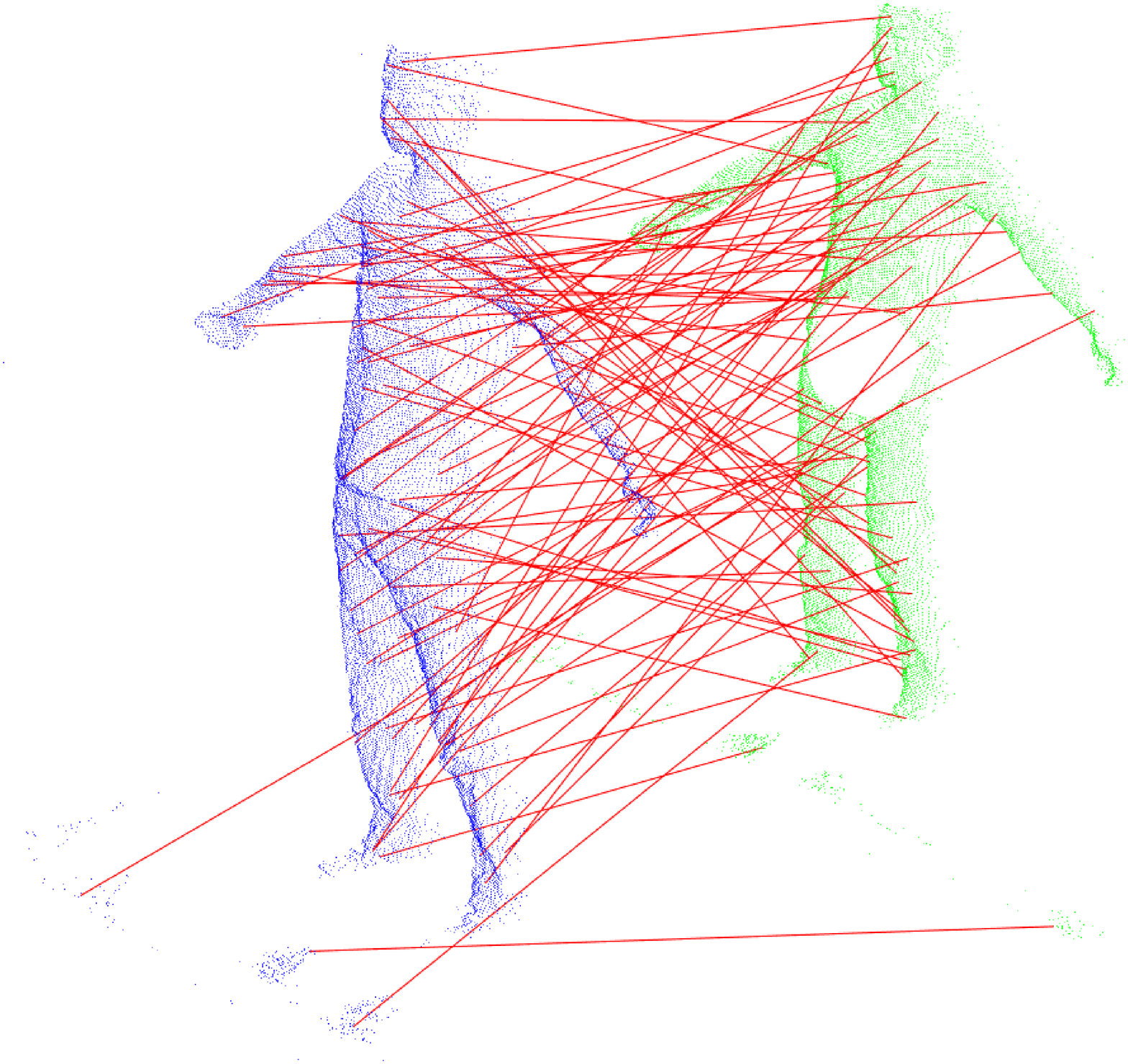}
}
\qquad
\subfloat[Matching by signature of histograms of orientations]{
\includegraphics[width=0.15\textwidth]{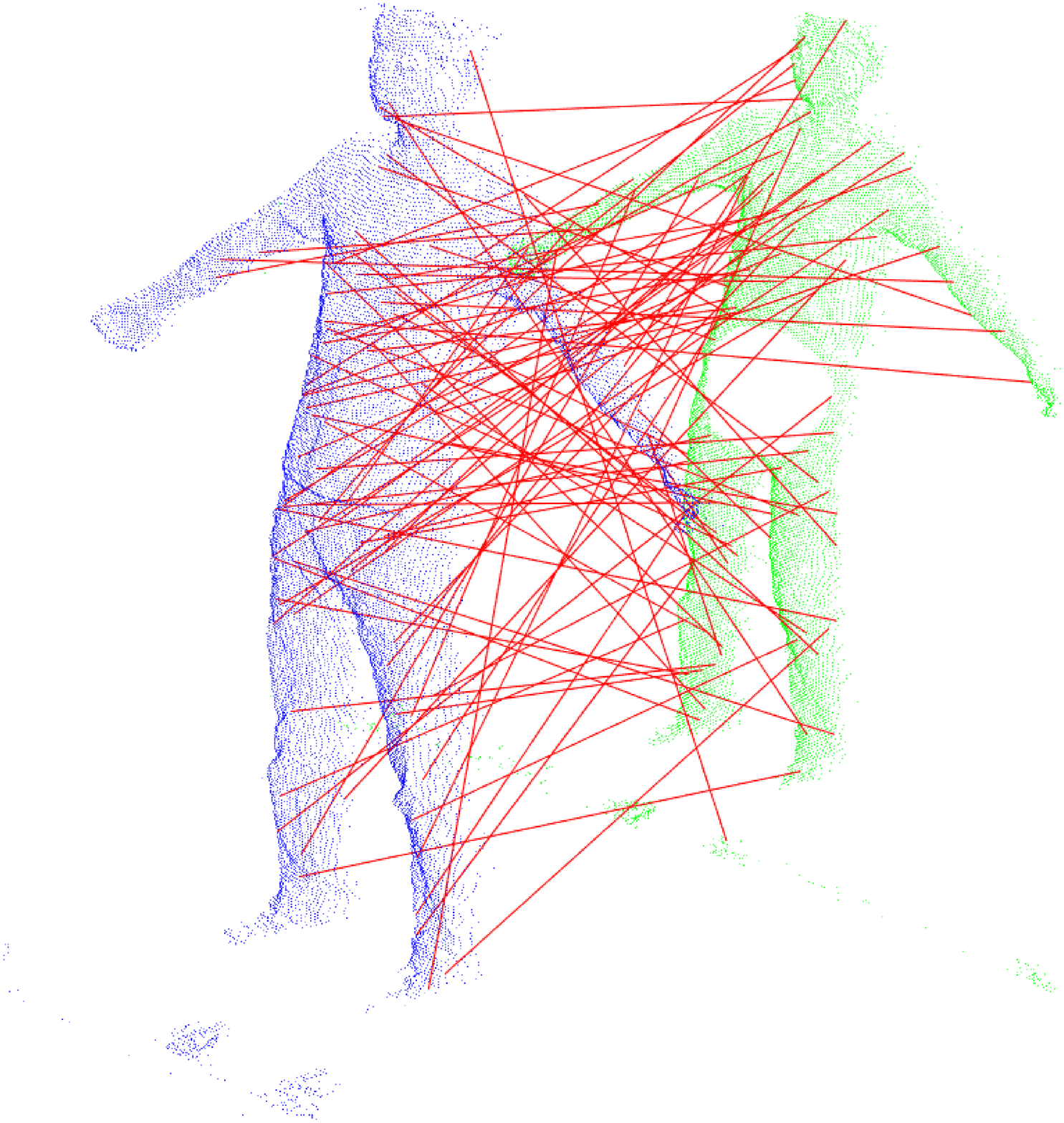}
}
\qquad
\subfloat[Matching with 3D shape context]{
\includegraphics[width=0.15\textwidth]{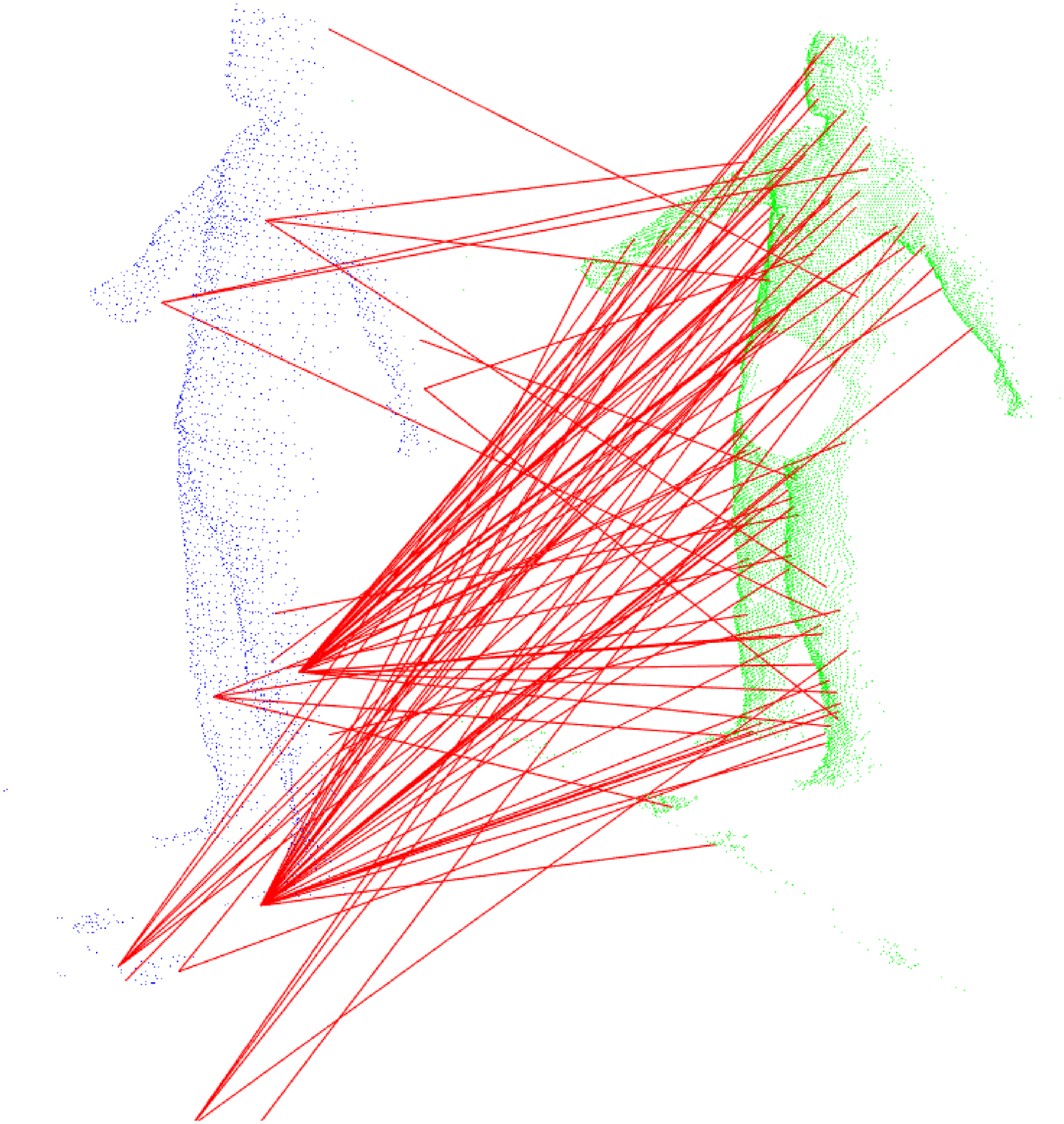}
}
\caption{\small Matching frames with local appearance based features: one percent of the matching pairs are
plotted.}
\label{track::appearance}
\vskip -0.25 true in
\end{figure}

\begin{figure}
\centering
\subfloat[Matching with heat kernel signature]{
\includegraphics[width=0.15\textwidth]{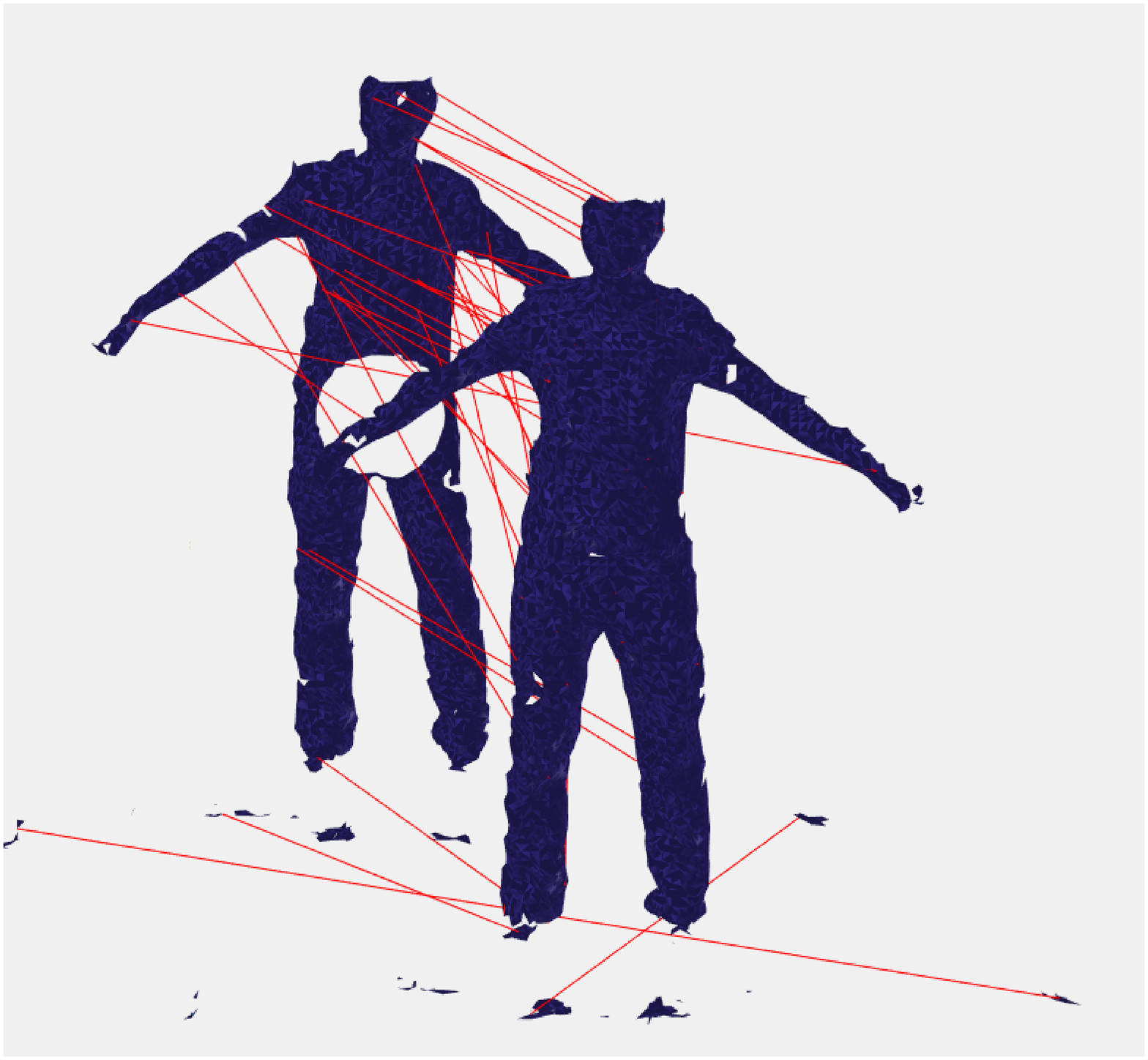}
}
\qquad
\subfloat[Matching by scale-invariant heat kernel signature]{
\includegraphics[width=0.15\textwidth]{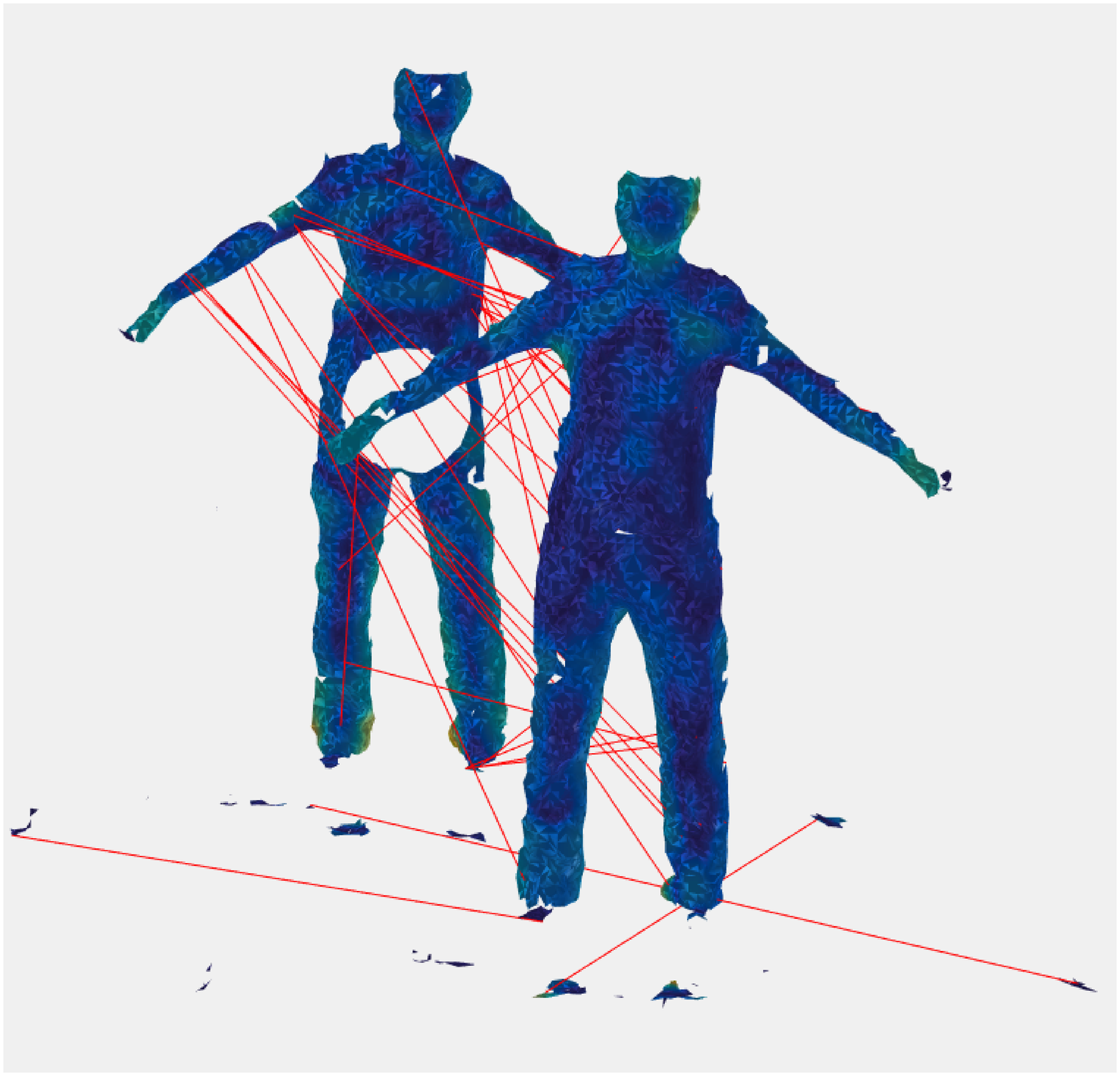}
}
\qquad
\subfloat[Matching  with wavelet kernel signature]{
\includegraphics[width=0.15\textwidth]{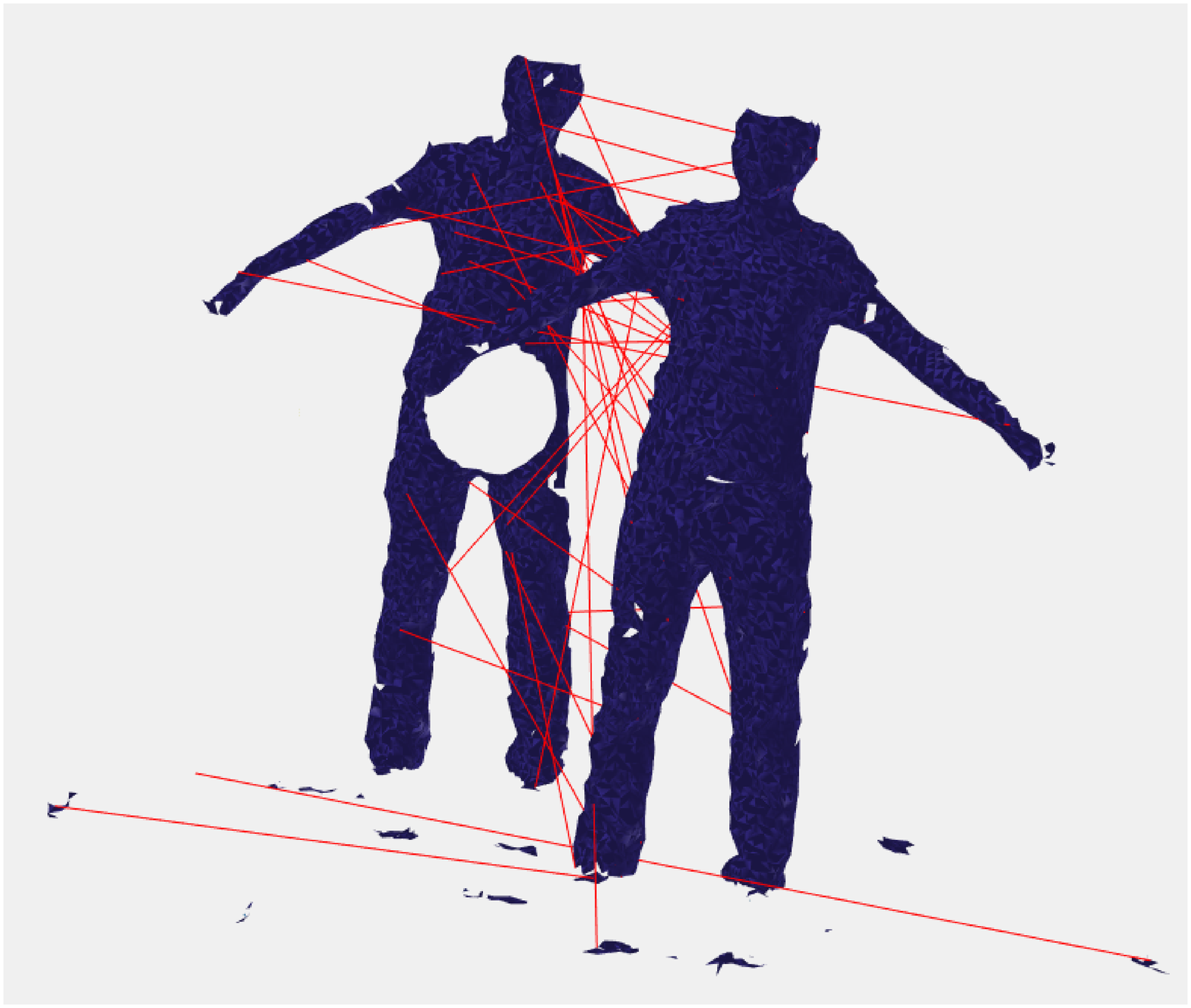}
}
\qquad
\subfloat[Matching with the proposed feature]{
\includegraphics[width=0.15\textwidth]{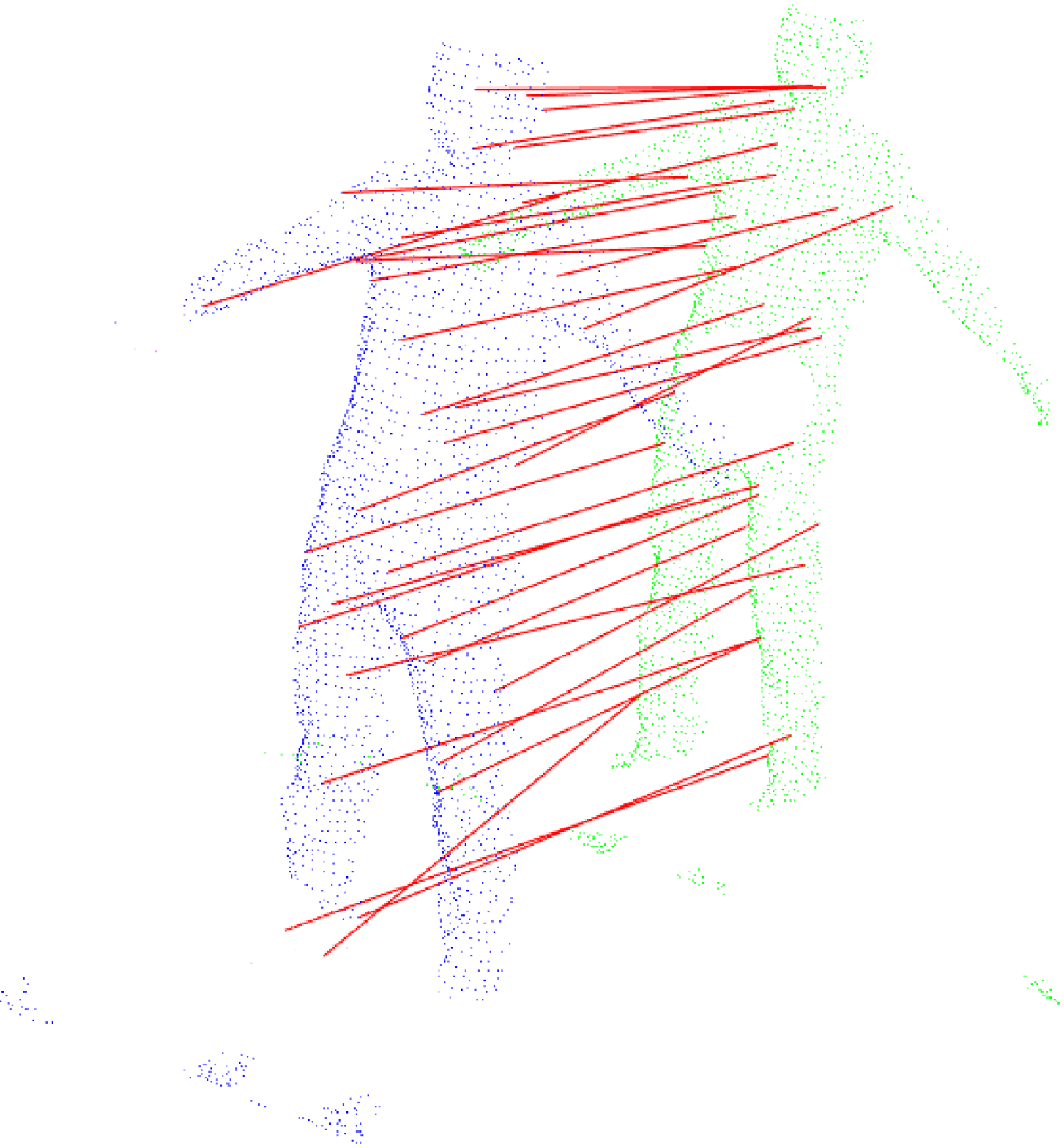}
}
\caption{\small Matching frames with features defined on surface mesh: one percent of the matching pairs are
plotted.}
\label{track::mesh}
\vskip -0.20 true in
\end{figure}

\begin{figure}
\centering
\subfloat[Tracked result 1]{
\includegraphics[width=0.15\textwidth]{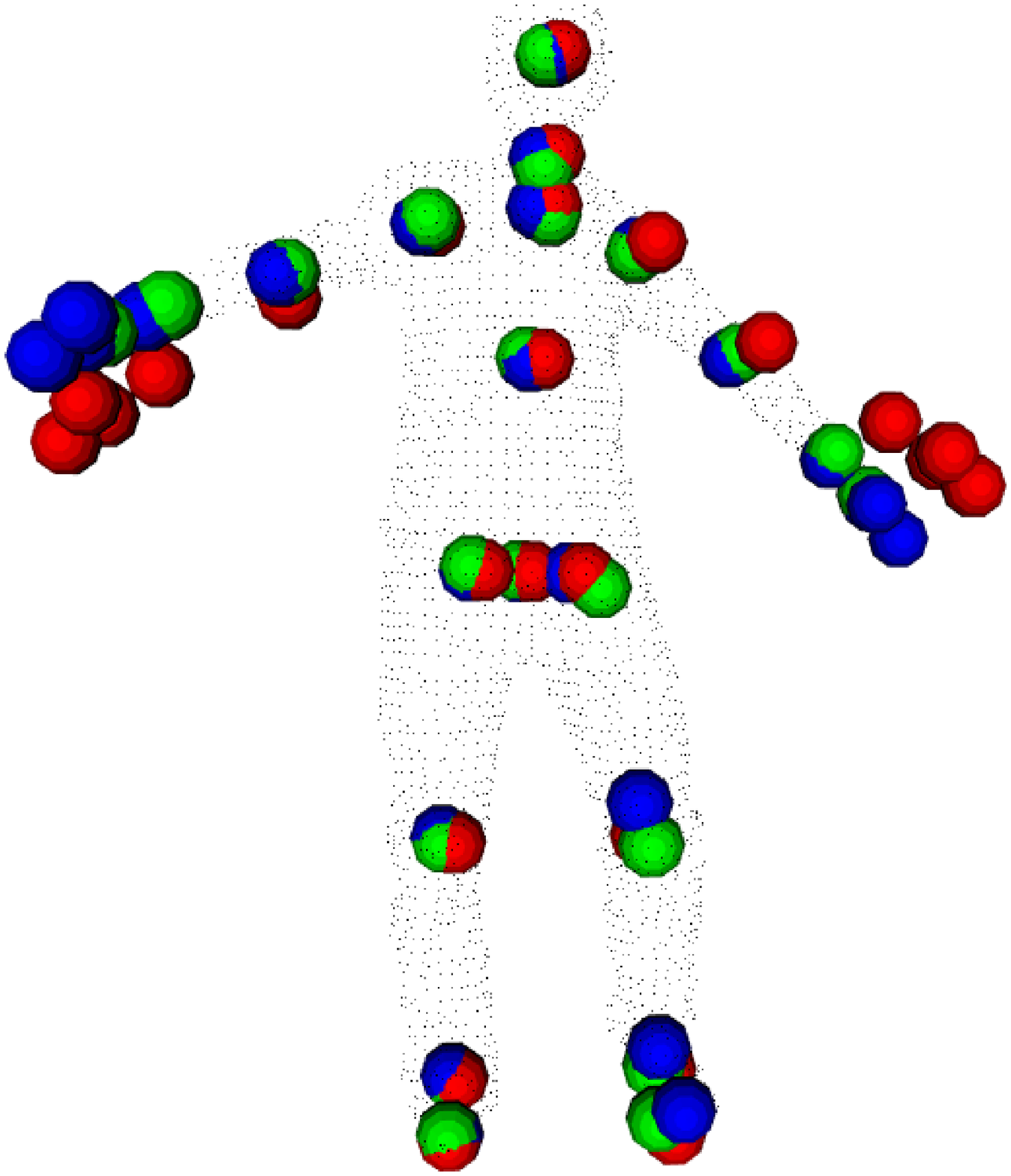}
}
\qquad
\subfloat[Tracked result 2]{
\includegraphics[width=0.15\textwidth]{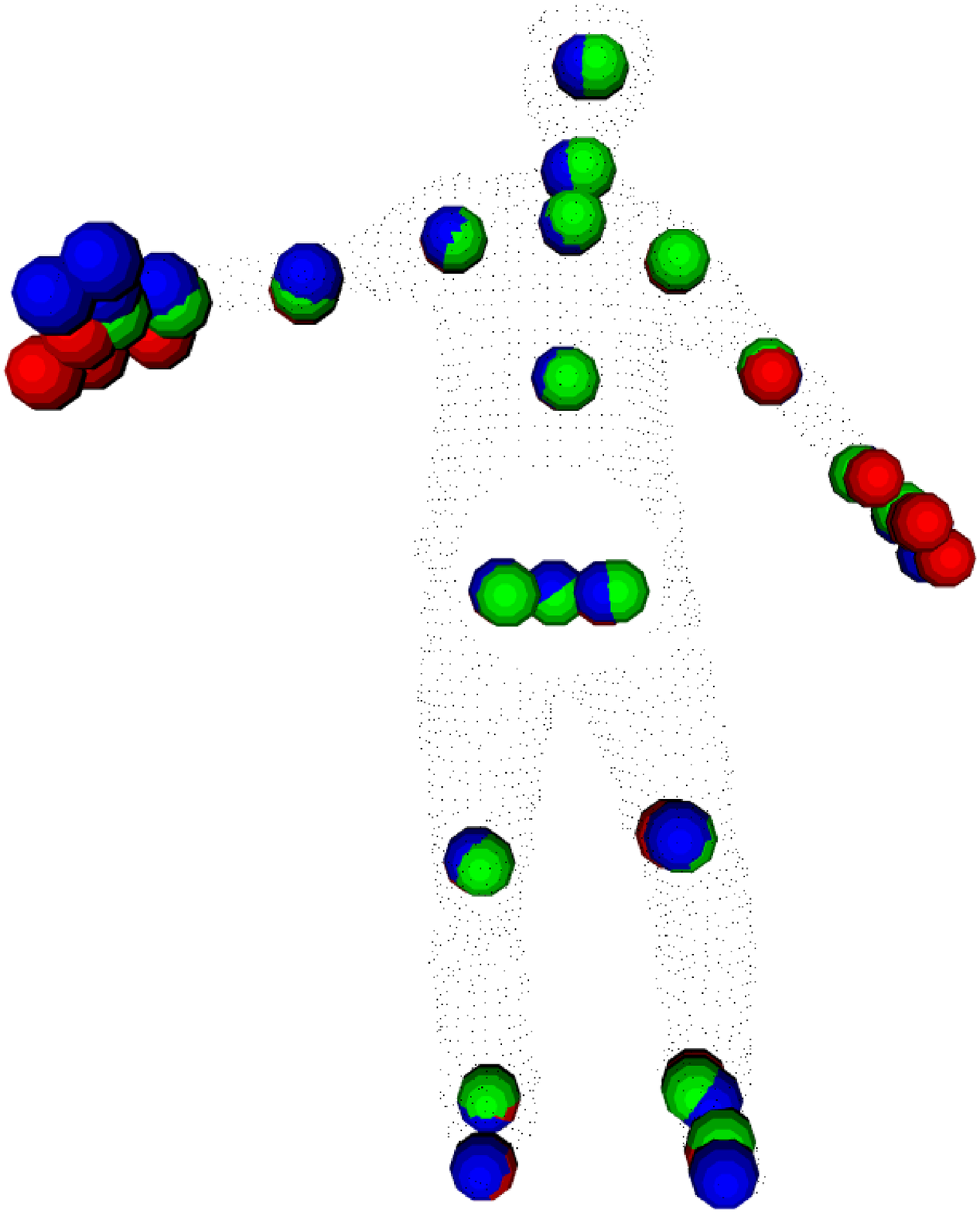}
}
\qquad
\subfloat[Tracked result 3]{
\includegraphics[width=0.15\textwidth]{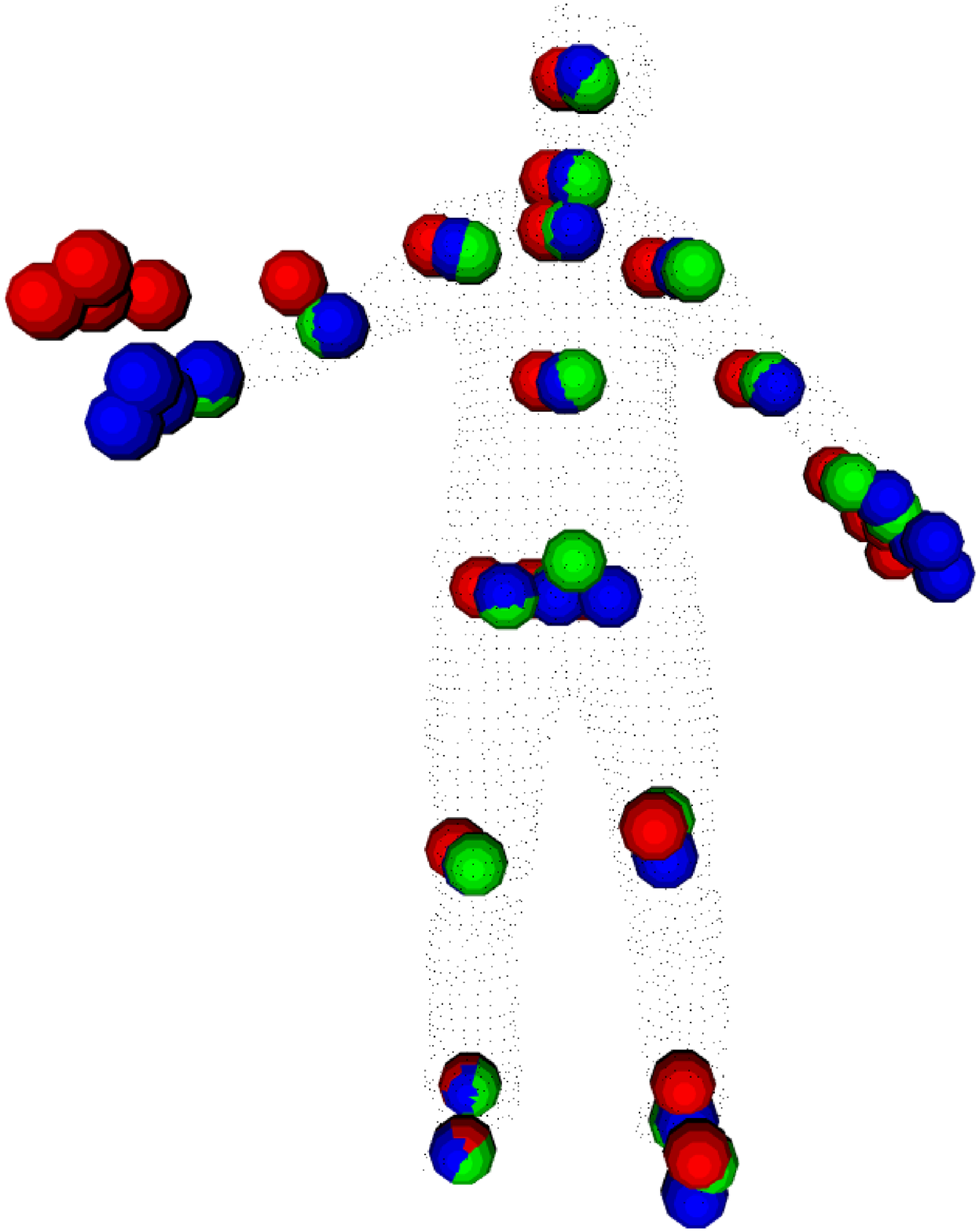}
}
\qquad
\subfloat[Tracked result 4]{
\includegraphics[width=0.15\textwidth]{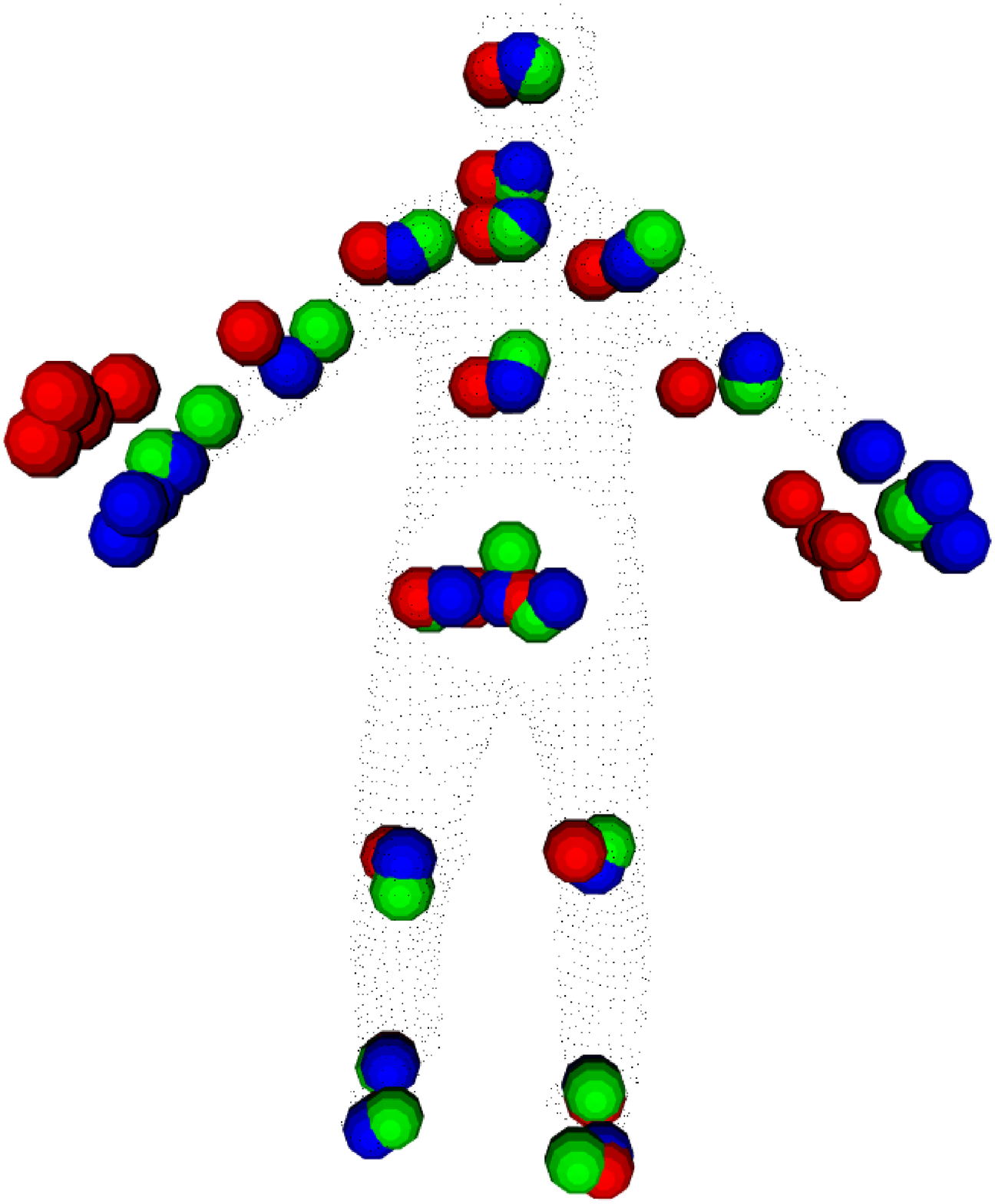}
}
\caption{\small Examples of MHT tracking results: blue dots are the ground truth, red dots
are the predicted joint locations, and green dots are the tracked joint locations.}
\label{tracking::mhtjoints}
\vskip -0.25 true in
\end{figure}

We test the proposed feature on a data collected with a Kinect V2 sensor. To use scaled motion
dynamics to predict the movements, we collect 200 data samples on one subject, with varying movement
patterns and speeds that vary from slow, normal to fast. The movement pattern includes upper and
lower body movements. For reference the skeletal joint locations estimated by the Kinect sensor on
unoccluded data are used as scaled motion dynamics training data.  All of the data samples were
captured from a front view.

To simulate transient occlusions, we collected 3000 unoccluded data samples of a human body under
different types of movements, along with estimated joint locations for ground truth evaluation, and
then simulated transient occlusions by removing data from randomly selected frames.  We placed
occlusions, of five different sizes (radii of 0.03 to 0.15 m), in the center of the body We
constructed the surface meshes from the corrupted data via triangulation, and then tried to track
the vertices across multiple frames.

The main variables for the geodesic feature are the radius to find the anchor points, the error
terms' upper bounds, and the tracking tree length.  We selected a 0.5 m radius empirically, but
this parameter can adapt to the subject's maximum height.  An upper bound on the anchors'
localization error was set at a constant 0.05 m.  The upper bound on surface point errors was chosen
as a linear function increasing with distance, since Kinect error increases with range.  The
tracking tree look-ahead depth was set to three.

For comparison, we also implemented other local feature-based tracking methods, including fast
point feature histogram (FPFH) \cite{tracking:fpfh}, signature of histograms of orientations (SHOT)
\cite{tracking:shot}, 3D shape context (SC) \cite{tracking:shape3d}, and rotational projection
statistics (ROPS) \cite{tracking:rops}. These local features are invariant under occlusions, and we
used them to associate neighboring frames in the tracking framework. We also implemented three local
features based on spectral shape analysis: heat kernel signature \cite{tracking:hks},
scale-invariant heat kernel signature \cite{tracking:sihks}, and wavelet kernel signature
\cite{tracking:wks}.  

\subsection{Qualitative Results}

First, we consider the matching of surface points across frames (under occlusion) with different
methods. A descriptor was extracted for each point in successive frames, and points were matched by
a nearest neighbor method. Figures \ref{track::appearance} and \ref{track::mesh} show that the best
matching typically occurs in the head, where the surface has stable curvature.  Although stable
curvatures also exist in the feet and hands, symmetries often cause the swapping of matches. For
other body areas, these features generally yield inaccurate matchings due to surface deformations
and homogeneous curvatures.

Figure \ref{tracking::mhtjoints} shows the joint locations estimated by our method.  While
occlusions will cause some mismatches, the method yields adequate estimates of joint locations.

\subsection{Quantitative Results}

\begin{figure}
\centering
\includegraphics[width=0.3\textwidth]{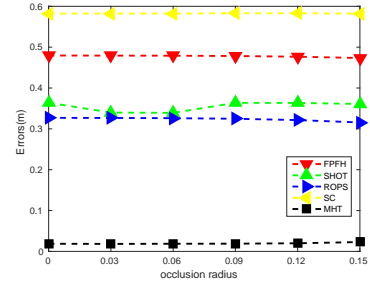}
\caption{\small Average tracking errors (all joints) under different occlusions. 
  Local features are invariant to occlusions, yielding almost constant errors
  under different occlusions.  The proposed framework also achieves almost constant
  tracking errors, implying that that it is similarly invariant to transient occlusions.}
\label{track::result1}
\vskip -0.15 true in
\end{figure}

\begin{figure}
\centering
\includegraphics[width=0.3\textwidth]{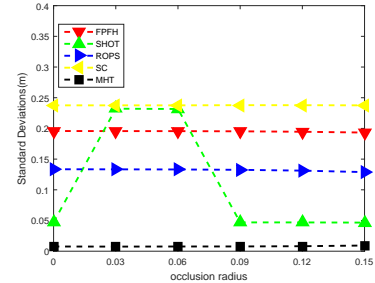}
\caption{\small Tracking error standard deviations for different occlusions.}
\label{track::sd1}
\vskip -0.2 true in
\end{figure}

\begin{figure}
\centering
\includegraphics[width=0.3\textwidth]{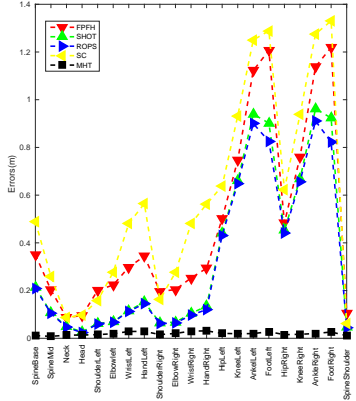}
\caption{\small Average joint tracking errors under occlusions. The local surface features only
  work well on stable, distinctive body regions (e.g. head, neck).  The proposed method has
  similar tracking error on every joint because it does not depend on local stable regions.}
\label{track::result2}
\vskip -0.2 true in
\end{figure}
\begin{figure}
\centering
\includegraphics[width=0.3\textwidth]{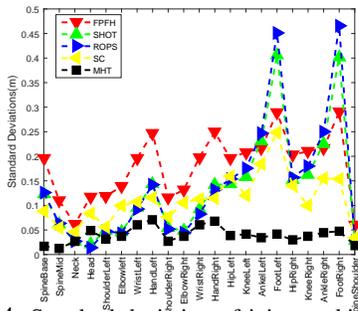}
\caption{\small Standard deviation of joint tracking errors.}
\label{track::sd2}
\vskip -0.25 true in
\end{figure}

Figures \ref{track::result1} and \ref{track::sd1} plot the tracking error and its standard deviation
for each occlusion type and joint. The tracking error for each occlusion is the average
joint tracking error:
    \[r_{or}=\frac{1}{21}\sum_i^{21}||\theta_i^{or}-\hat{\theta}_i||\]
where $\theta_i^{or}$ denotes the $i^{th}$ estimated joint location under occlusion $or$ (ranging
from 0 to 0.15 m), and $\hat{\theta}_i$ denotes the ground truth.  Figure \ref{track::result2} plots
the average tracking error for each joint under all types of occlusions: $ r_i=(1/6)
\sum_{or=0}^{0.15}||\theta_i^{or}-\hat{\theta}_i||$. Figure \ref{track::sd2} shows the standard
deviation of this error.  These results show that our method achieves invariance to transient
occlusions like local feature method, while being robust to homogeneous surface regions and
deformations.  Thus it is more accurate under occlusions.

Besides, it takes about 0.72 seconds to create a mesh with 15549 nodes and 36185 edges on one core of
Intel i7-6700 CPU, and about 0.23 seconds to compute the gedoesic feature with five anchor
nodes.
\section{Conclusions} \label{tracking:conclusions}

This work introduced a method to use RGB-D data to track the human body under transient occlusions.
The difficulty mainly lies in data association under surface deformations, transient occlusions, and
homogeneous surface appearances. Tracking with appearance-based local features are generally robust
to transient occlusions, but the tracking is inaccurate due to regions of homogeneous texture,
intensity and curvature on human body, and tracking with local features based on spectral shape
analysis has similar problems. We propose a solution by designing a geodesic feature robust to
surface deformations and homogeneous regions and improving its robustness to transient occlusions by
using the multi-hypothesis tracking framework. The result shows that this solution achieves
invariance to transient occlusions as well as robustness to homogeneous surface appearances.

Future work will aim to combine the geodesic-feature with appearance-based local features to achieve
greater invariance under transient occlusions and greater tracking accuracy.


\end{document}